\documentclass{article}
\usepackage{iclr2027_conference,times}

\usepackage[table]{xcolor}
\usepackage{colortbl}
\usepackage{graphicx}
\usepackage{booktabs}
\usepackage{amsmath,amssymb,amsfonts}
\usepackage{multirow}
\usepackage{pifont}
\usepackage{fancyvrb}
\usepackage{url}
\usepackage{wrapfig}
\usepackage[inline]{enumitem}
\usepackage{caption}
\usepackage{microtype}
\usepackage{titletoc}
\usepackage[ruled,vlined]{algorithm2e}

\usepackage{hyperref}
\hypersetup{
  colorlinks,
  citecolor=[rgb]{0.10,0.50,0.15},
  linkcolor=[rgb]{0.75,0.20,0.15},
  urlcolor=[rgb]{0.80,0.15,0.55}
}

\usepackage[capitalize]{cleveref}
\crefname{section}{Section}{Sections}

\newcommand{\method}{FGPO}
\newcommand{\methodlong}{Full-Group Policy Optimization}
\newcommand{\bb}[1]{\textbf{#1}}

\newcommand{\best}[1]{\textbf{#1}}

\definecolor{grouprow}{RGB}{234,236,247}
\definecolor{gtgreen}{RGB}{0,120,44}
\definecolor{wrongred}{RGB}{200,30,30}
\definecolor{venuegray}{HTML}{7F807F}
\definecolor{bandgray}{HTML}{F2F3F8}
\definecolor{codegray}{HTML}{58585A}
\definecolor{codebg}{HTML}{FAFAFA}

\newcommand{\secref}[1]{Section~\ref{#1}}

\title{Why Sample What You Can Enumerate?\\
Exact Policy Optimization for Genomic Tool Selection}

\author{
\textbf{Haoyue Liu}\textsuperscript{1,3}
\quad
\textbf{Xiaoyu Ma}\textsuperscript{1}
\quad
\textbf{Ye Chen}\textsuperscript{2}
\quad
\textbf{Zhichao Wang}\textsuperscript{1}
\quad
\textbf{Xiaoying Tang}\textsuperscript{1,3,\ensuremath{\dagger}}
\\[0.6em]
\textsuperscript{1}
School of Science and Engineering,
The Chinese University of Hong Kong, Shenzhen 518172, China
\\
\textsuperscript{2}
XJTU-POLIMI Joint School,
Xi'an Jiaotong University, Xi'an 710049, China
\\
\textsuperscript{3}
Shenzhen Future Network of Intelligence Institute (FNii-Shenzhen)
}

\iclrfinalcopy 

\begin{document}

\maketitle

\begin{abstract}
Reinforcement learning over a frozen reasoner has become a common recipe for teaching a
policy which external tools to invoke. We show that this recipe becomes structurally
mismatched in specialist scientific settings where the complete tool-subset space is
enumerable. There, a small set of recurring computational capabilities covers the domain,
so the space of tool subsets is combinatorial yet small enough to enumerate, and GRPO
still estimates an action expectation from a handful of sampled rollouts. Worse, the
approximation degrades as training succeeds: as the policy concentrates on preferred
subsets it resamples them, sampled rewards collide, and the group-normalized advantage
vanishes. On genomic reasoning the fraction of questions yielding no reward signal rises
from $0.2\%$ under a uniform reference policy to $20.8\%$ after GRPO training. As a remedy,
we introduce \method{} (\methodlong{}), which (1) scores every tool subset and optimizes
the exact action expectation, so each update sees the complete action space, and (2)
precomputes the reward of each question--subset pair into an exhaustive table, removing
frozen-reasoner calls from the training loop entirely. Across five frozen reasoners and
three genomic benchmarks, \method{} outperforms GRPO in all $15$ settings by $6.75$ points
on average and up to $14.20$, while a standard on-demand GRPO schedule would
require $2.4\times$ as many frozen-reasoner reward evaluations and, on GenomeQA, \method{} cuts invoked tools per question
from $2.36$ to $1.40$.
\end{abstract}

\section{Introduction}
\label{sec:intro}

Large language models can interpret specialized scientific questions but cannot reliably
perform the precise computations many of them require. Genomic reasoning is a representative
case: an LLM understands promoters, transcription-factor binding and splice sites, yet
answering such questions demands explicit computation over raw nucleotide sequences: motif
scanning, splice-site scoring, composition analysis \citep{jin2024genegpt}. External tools
supply exactly these capabilities, and a productive line of work therefore trains a policy
to select which tools a frozen reasoner should receive, optimizing that policy with
reinforcement learning: VisTA for visual tools \citep{huang2025visualtoolagent}, AuTAgent for audio tools
\citep{tong2026autagent}, and reward-shaped variants \citep{qian2026toolrl,jin2025search},
almost always with GRPO \citep{shao2024deepseekmath}.

\begin{figure}[t]
\centering
\includegraphics[width=\linewidth]{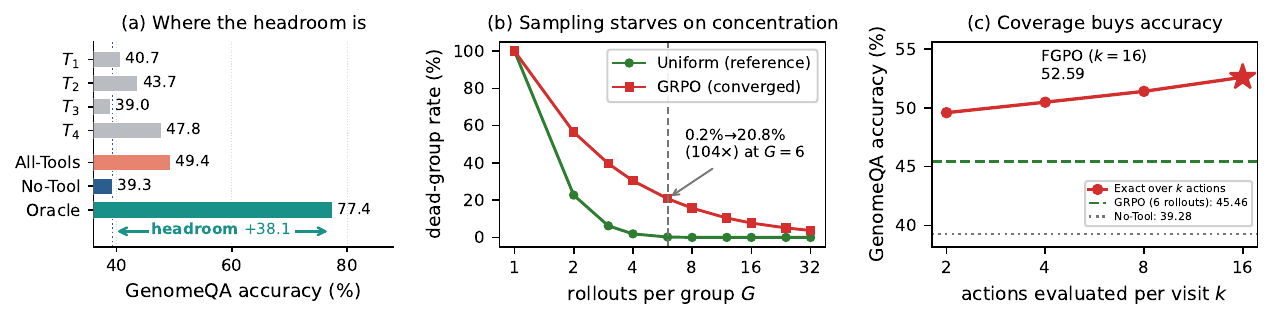}
\caption{\textbf{Why sample when enumeration is possible?}
(a) Individual tools, All-Tools, No-Tool, and a per-question oracle on the $3{,}590$
GenomeQA test questions; the oracle reveals $38.1$ points of recoverable headroom.
(b) Closed-form probability that all $G$ rollouts receive identical reward, forming a
\emph{dead} group with zero reward advantage.
(c) Accuracy when the same training construction observes only $k$ uniformly sampled
actions per visit.}
\label{fig:teaser}
\end{figure}

Despite its success elsewhere, this recipe carries an unavoidable drawback in the
specialist regime studied here. Because a domain reuses a small set of recurring
capabilities, a compact library covers it; and even where the global registry is large, the
per-query active set is small: on BFCL, an adaptive shortlist of about seven tools
($7.4\pm2.5$) from a $370$-tool registry retains $90.3\%$ correct-tool coverage, nearly
matching $90.8\%$ with fifty \citep{repantis2026many}.
Our four genomic tools admit only $2^4{=}16$ subsets, making the action expectation exactly computable, yet GRPO still approximates it through sampled rollouts. Selection genuinely matters at this scale.
As illustrated in Figure~\ref{fig:teaser}a, the frozen reasoner scores $39.28\%$ without
tools while a per-question oracle over the same library reaches $77.41\%$, leaving $38.1$
points of recoverable headroom that invoking all tools ($49.39\%$) does not capture. But
sampling that space is not merely wasteful: it degrades precisely as optimization
succeeds. A group teaches the policy nothing when its sampled rewards coincide, since every
normalized advantage is then zero, and a concentrating policy resamples the same subsets.
Figure~\ref{fig:teaser}b shows this \emph{dead-group} rate climbing from $0.2\%$ under a
uniform reference policy to $20.8\%$ after GRPO training, reaching $79.6\%$ under the
differential reward GRPO trains with; Figure~\ref{fig:teaser}c shows accuracy rising
monotonically as the optimizer is shown more of the action space. This motivates the
central question of this paper:
\begin{center}
\emph{Why sample an expectation that can be computed exactly?}
\end{center}

As a remedy, we introduce \method{} (\methodlong{}), an exact policy optimization
framework for enumerable tool-selection spaces. \method{} incorporates (1) an exact
objective that scores every tool subset and optimizes the complete action expectation, so
each update sees the whole action space rather than a sample of it, and (2) an exhaustive
reward table that precomputes each question--subset pair once, removing frozen-reasoner
calls from the training loop entirely. For autoregressive LLM policies we further use
per-token length-normalized candidate scoring and entropy regularization to obtain a
practical realization of this objective.

Our contributions are summarized as follows:
\begin{itemize}[leftmargin=1.6em,itemsep=1pt,topsep=2pt]

\item We formulate tool-augmented genomic reasoning as query-dependent combinatorial
tool-subset selection and reveal $38.1$ points of per-question oracle headroom over the
unaided reasoner, demonstrating the importance of selecting specialist capabilities
correctly.

\item We identify a mismatch between sampled policy optimization and enumerable
tool-selection spaces. On GenomeQA, the dead-group rate rises from $0.2\%$ to $20.8\%$
after GRPO training, with similar rates of $17.4$--$17.8\%$ on two additional benchmarks;
controlled $k$-subset experiments further show that accuracy improves monotonically as the
same training construction is given broader uniform action coverage.

\item We introduce \method{}, which replaces sampled optimization with exact optimization
over all enumerable tool subsets. Across five reasoners and three genomic benchmarks,
\method{} outperforms GRPO in all $15$ settings by $6.75$ points on average and up to
$14.20$ points, while a standard on-demand GRPO schedule would require $2.4\times$ more
frozen-reasoner reward evaluations
and on GenomeQA \method{} reduces average invoked tools from $2.36$ to $1.40$.
\end{itemize}

\section{\method{}: Exact Policy Optimization over Enumerable Tool Subsets}
\label{sec:method}

\begin{figure}[t]
\centering
\includegraphics[width=0.7\linewidth]{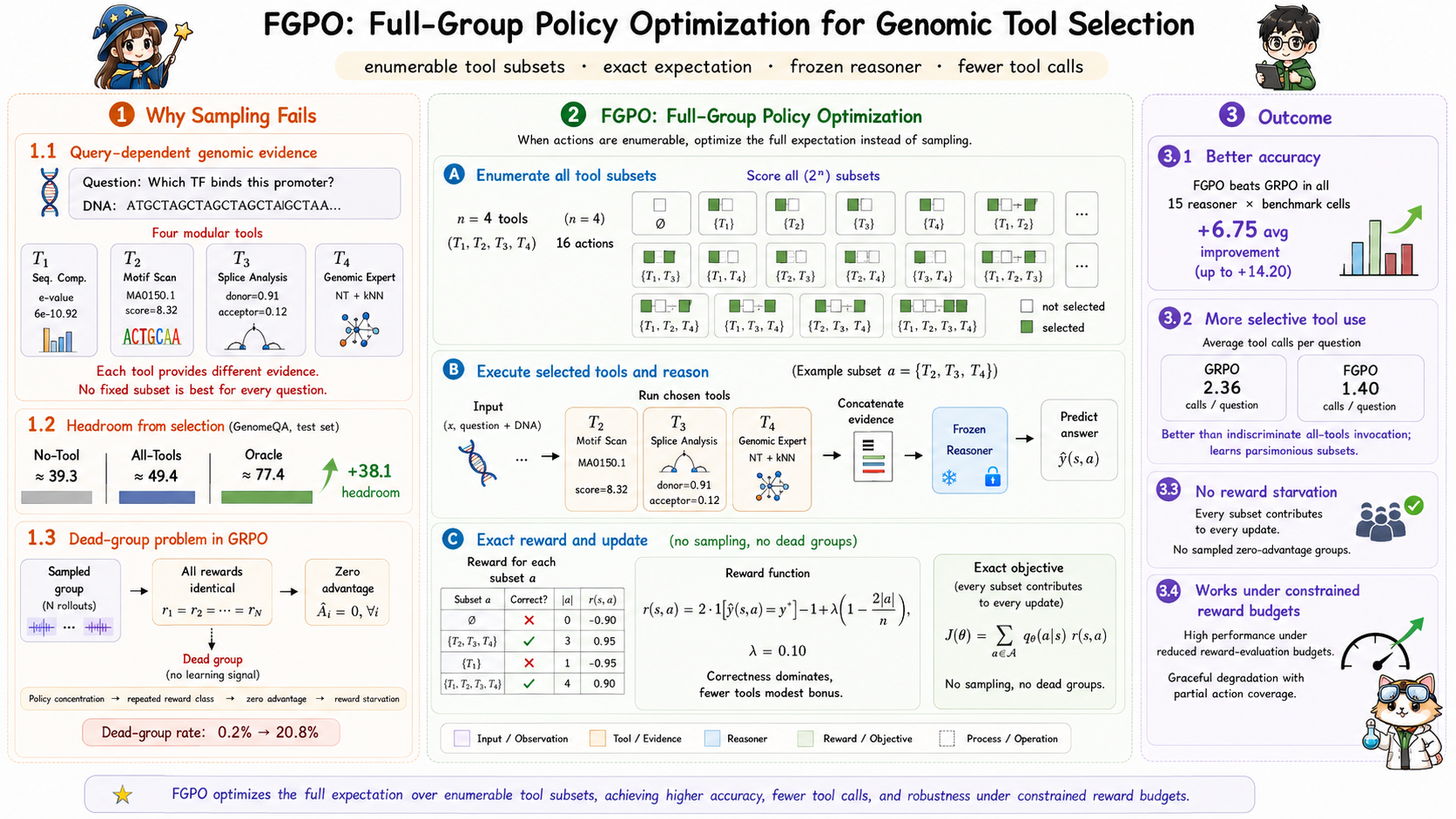}
\caption{\textbf{Overview of \method{}.} A trainable LoRA policy scores all $2^{n}$ tool
subsets for a question; each subset's tools are executed and their evidence handed to a
frozen reasoner, and the resulting rewards, precomputed once into an exhaustive
table, enter the exact objective $J(\theta)=\sum_a q_\theta(a|s)\,r(s,a)$, so every
subset contributes to every update. The sampled alternative draws $G$ rollouts instead,
leaving $20.8\%$ of questions at convergence with a single reward class under
Eq.~\ref{eq:reward} ($79.6\%$ under the sampled baseline's coarser reward) and hence zero
advantage.}
\label{fig:overview}
\end{figure}

This section develops \method{} as exact policy optimization for LLM tool selection: the
objective over all $2^{n}$ tool subsets (\secref{sec:method:objective}), two choices its LLM
instantiation needs plus a controlled candidate-scoring evaluation
(\secref{sec:method:llm}), and an exhaustive reward table that precomputes $r(s,a)$ once
(\secref{sec:method:table}). Algorithm~\ref{alg:fgpo} in \secref{app:algorithm} states the
whole procedure.

\subsection{Setting}
\label{sec:method:setting}
Let $\mathcal{D}$ denote the training-question distribution. A frozen reasoner $R$ answers
multiple-choice genomic questions $s\sim\mathcal{D}$, optionally given the output of a
subset $a \subseteq \{T_1,\dots,T_n\}$ of tools run on the question's sequence. The tools are
independent analyses of that sequence, none consuming another's output, so a subset fully
specifies the execution.
A policy $q_\theta(a\,|\,s)$, with $\theta$ the parameters of a LoRA adapter, selects the
subset; the action space is
$\mathcal{A}=2^{\{T_1,\dots,T_n\}}$ with $|\mathcal{A}|=2^{n}$ ($n{=}4$ in all main experiments,
$16$ actions; Table~\ref{tab:tools} lists the library). With $y^{\star}$ the gold option and $\hat{y}(s,a)$ the reasoner's answer given
the evidence of subset $a$, the reward scores correctness and adds a parsimony tie-break an
order of magnitude smaller:
\begin{equation}
r(s,a)\;=\;\underbrace{2\cdot\mathbf{1}\!\left[\hat{y}(s,a)=y^{\star}\right]-1}_{\pm1}
\;+\;\lambda\left(1-\frac{2|a|}{n}\right),
\qquad \lambda=0.10,
\label{eq:reward}
\end{equation}
\begin{wraptable}{r}{0.50\linewidth}
\vspace{6pt}
\caption{\textbf{The modular tool library.} All tools are frozen; $T_4$ uses source-training
$k$NN. Solo worth is shown in Figure~\ref{fig:teaser}a.}
\label{tab:tools}
\vspace{2pt}
{\footnotesize
\setlength{\tabcolsep}{2.5pt}
\begin{tabular}{@{}clc@{}}
\toprule
ID & Tool Module & Source\\
\midrule
$T_1$ & Seq.\ Composition & \citet{cock2009biopython}\\
$T_2$ & Motif Scan & \citet{castro2022jaspar}\\
$T_3$ & Splice Analysis & \citet{yeo2003maximum}\\
$T_4$ & Genomic Expert & \citet{dalla2025nucleotide}\\
\bottomrule
\end{tabular}}
\vspace{-8pt}
\end{wraptable}

so correctness always dominates and, among subsets that agree on it, fewer tools score
higher. We follow the \emph{setting} of \citet{tong2026autagent} (frozen reasoner,
tool-subset action space, RL-trained selector) but not its differential reward
(\secref{sec:exp:setup}). The policy is a 7B LLM with a LoRA adapter \citep{hu2021lora}
that emits the subset as a short indexed string over \emph{anonymous} tool slots, with no tool
names or descriptions, so any routing it learns comes from reward rather than text
(Appendix~\ref{app:prompts}).

\subsection{The exact objective}
\label{sec:method:objective}
With a frozen, greedily decoded reasoner we treat $r(s,a)$ as effectively deterministic
(an independent live pipeline agrees with the cache within $0.05$ points,
\secref{app:full}). The \emph{action} expectation in the policy-gradient objective and its
gradient is then a finite sum, evaluated exactly; the expectation over questions is
minibatched as usual,
\begin{equation}
J(\theta) \;=\; \mathbb{E}_{s\sim\mathcal{D}} \sum_{a\in\mathcal{A}} q_\theta(a\,|\,s)\, r(s,a),
\qquad
\nabla_\theta J \;=\; \mathbb{E}_{s} \sum_{a\in\mathcal{A}} r(s,a)\, \nabla_\theta\, q_\theta(a\,|\,s),
\label{eq:exact}
\end{equation}
computable without sampling whenever $2^{n}$ candidate evaluations per visit are
affordable. \method{} optimizes Eq.~\ref{eq:exact} directly over the tool-subset space,
connecting to expected and all-action policy gradients
\citep{ciosek2020expected,asadi2017mean} in the contextual-bandit setting
\citep{langford2007epoch}. Contrast
GRPO \citep{shao2024deepseekmath}, which samples $G$ rollouts from the autoregressive
generation policy. Let $p_\theta^{\mathrm{gen}}(a\,|\,s)$ denote the categorical
distribution over parsed tool subsets induced by that sampler; then
$a_1,\dots,a_G\sim p_\theta^{\mathrm{gen}}(\cdot\,|\,s)$,
and GRPO steps on group-normalized advantages, where $r$ is
whatever reward that arm trains on, the differential reward in our experiments
(\secref{sec:exp:setup}):
\begin{equation}
\hat{A}_i \;=\; \frac{r(s,a_i)-\mu}{\sigma+\epsilon},
\qquad
\mu=\frac{1}{G}\sum_{j=1}^{G} r(s,a_j),
\qquad
\sigma^{2}=\frac{1}{G}\sum_{j=1}^{G}\bigl(r(s,a_j)-\mu\bigr)^{2}.
\label{eq:grpo}
\end{equation}
with $\epsilon>0$ for numerical stability. Two limitations follow. Whenever the $G$ drawn
rewards coincide every numerator is zero, so all advantages vanish regardless of
$\epsilon$, a \emph{dead} group, which \secref{sec:exp:mechanism} shows reaches
$104\times$--$178\times$ the uniform-reference rate. And because the group is drawn from
$p_\theta^{\mathrm{gen}}$ itself, coverage shrinks exactly as the sampling policy
concentrates, which is what optimization produces. Eq.~\ref{eq:exact} evaluates every subset
without sampled groups: concentration can shrink reward gradients through $q_\theta$, but
no action is omitted by sampling (Appendix~\ref{app:theory}).

\begin{figure}[t]
\centering
\includegraphics[width=\linewidth]{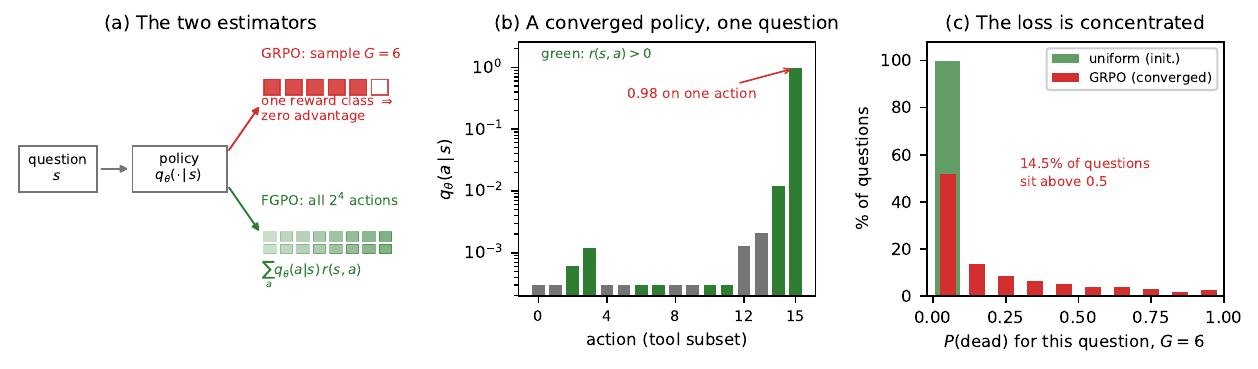}
\caption{\textbf{What each estimator sees.} (a) The two estimators on the same state.
(b) The converged GRPO policy's exported distribution over the $16$ subsets for one
GenomeQA question; green marks positive reward. (c) Per-question dead probability at
\(G{=}6\) over all \(3{,}590\) test questions, with reward classes from Eq.~\ref{eq:reward}; the mean is \(20.8\%\), as reported in \secref{sec:exp:mechanism}.}
\label{fig:method}
\end{figure}

Figure~\ref{fig:method} makes this concrete on real data: the converged GRPO policy puts
$98\%$ of its mass on a single subset (panel~b), and the per-question dead probability is
heavily skewed (panel~c), so sampling goes blind precisely where the policy has already
committed.

\subsection{Instantiating the exact estimator on an LLM policy}
\label{sec:method:llm}
Two choices separate Eq.~\ref{eq:exact} from a working method on an LLM policy, each
motivated by an observed failure; we additionally use a controlled candidate-scoring
evaluation to compare trained policies under an identical decoding rule.

\textbf{(i) Per-token length normalization.}
The $2^{n}$ actions are strings of $5$--$13$ tokens and the empty subset is the shortest,
so a softmax over raw sequence log-probabilities embeds a prior toward invoking
nothing, a bias absent when actions are abstract indices. With $y_a$ the token string
encoding $a$ and $\pi_\theta$ the policy's next-token distribution, we score candidates by
their per-token mean. Here $q_\theta$ denotes the candidate-normalized categorical
distribution used by \method{}; it is distinct from the generation-induced
$p_\theta^{\mathrm{gen}}$ used by the GRPO sampler in Eq.~\ref{eq:grpo}:
\begin{equation}
\ell_\theta(a\,|\,s)=\frac{1}{|y_a|}\sum_{t=1}^{|y_a|}\log \pi_\theta\!\left(y_{a,t}\,\middle|\,s,\,y_{a,<t}\right),
\qquad
q_\theta(a\,|\,s)=\frac{\exp \ell_\theta(a\,|\,s)}{\sum_{a'\in\mathcal{A}}\exp \ell_\theta(a'\,|\,s)}.
\label{eq:score}
\end{equation}

\textbf{(ii) Entropy regularization.}
To preserve per-question discrimination we add an entropy regularizer and optimize
\begin{equation}
J_\beta(\theta)=J(\theta)+\beta\,\mathbb{E}_{s\sim\mathcal{D}}\,H\!\left(q_\theta(\cdot\,|\,s)\right),
\qquad H(q)=-\!\sum_{a\in\mathcal{A}}\! q(a)\log q(a),
\qquad \beta=0.03,
\label{eq:entropy}
\end{equation}
which encourages broader candidate probabilities during optimization.

\textbf{(iii) A controlled candidate-scoring evaluation.}
Training already scores every candidate, so a controlled evaluation can reuse that score
rather than generating free-form text:
\begin{equation}
\hat{a}(s)=\operatorname*{arg\,max}_{a\in\mathcal{A}}\; \ell_\theta(a\,|\,s).
\label{eq:decode}
\end{equation}
Format failure is removed by construction rather than by parser engineering. The rule is
available to any autoregressive policy, so we apply it to the baselines too: scoring all
$16$ candidates instead of generating freely moves GRPO $45.46\!\to\!45.13$, SFT
$42.17\!\to\!39.30$ and \method{} $52.62\!\to\!51.84$, leaving \method{} ahead by
$6.71$ points under an inference rule identical across arms. Table~\ref{tab:main} reports free generation throughout.

\subsection{The exhaustive reward table}
\label{sec:method:table}
We precompute $r(s,a)$ for all $2^{n}$ actions of every training question. The table is memoization ( \method{} could evaluate rewards on demand) but it makes the
economics explicit: building the table takes $2^{n}\,|\mathcal{D}| = 16 \times 2{,}002 =
32{,}032$ reasoner calls, whereas the reported GRPO schedule ($200$ steps $\times\,384$
rollouts) entails $76{,}800$ reward evaluations, a $2.4\times$ difference for a standard
on-demand implementation. In our controlled experiments both arms read the same cached
outcomes, so this cost is an accounting fact about the schedules rather than a difference
between the runs we report; and everything downstream becomes a lookup with no further frozen-reasoner calls. Under a constrained frozen-reasoner budget, evaluating a uniformly drawn $k$-subset of actions
with $q_\theta$ renormalized on that subset degrades gracefully: at $k{=}2$, one third of
GRPO's budget, accuracy still leads by $4$ points (\secref{sec:exp:kcurve}).

\section{Experiments}
\label{sec:exp}

We organize the evaluation around five questions, each answered by the correspondingly
numbered subsection. \textbf{Q1:} Does \method{} outperform sampled and offline baselines
under matched data, prompts and adapters, and do the learned selections transfer across
frozen reasoners? \textbf{Q2:} What selection behaviour does each objective actually
produce? \textbf{Q3:} \emph{Why} does the sampled estimator underperform, and does the
mechanism replicate? \textbf{Q4:} Does action-space coverage drive the gain?
\textbf{Q5:} What does the parsimony term in the reward buy? Additional
numerical results, including the full transfer grid, paired significance tests, ceiling
analyses, dead-group grids, library ablations, GRPO/DPO sweeps and cost accounting, are in
Appendices~\ref{app:full}--\ref{app:mechanism}.

\subsection{Experiment Setup}
\label{sec:exp:setup}
\textbf{Benchmarks and tools.} Three multiple-choice genomic QA suites on open corpora:
\emph{GenomeQA} \citep{long2026genomeqa} ($3{,}590$ test /
$492$ dev) and two held-out cross-domain suites,
\emph{GenBench-X} and \emph{BM4} ($1{,}000$ each)
\citep{grevsova2023genomic,de2022deepstarr}. The $2{,}002$ training questions come
from the Nucleotide Transformer downstream tasks \citep{dalla2025nucleotide}; the
selector is trained on none of the three evaluation benchmarks. The $n{=}4$ tools span sequence composition ($T_1$), motif
scanning ($T_2$), splice-site analysis ($T_3$), and a $k$NN predictor over frozen
Nucleotide-Transformer embeddings ($T_4$).
\textbf{Policy, baselines and reasoners.} The policy is Qwen2.5-7B-Instruct
\citep{qwen2025qwen25technicalreport} with a rank-16 LoRA; GRPO, DPO and SFT use the official TRL implementations
\citep{vonwerra2020trl} on identical prompts, data and adapters, GRPO and SFT reported at their
development-selected checkpoint and DPO at the best cell of its sweep, with DPO given the
\emph{true} argmax from the exhaustive table as its chosen response. The five frozen reasoners are Qwen2.5-1.5B/7B \citep{qwen2025qwen25technicalreport}, Qwen3-8B
\citep{yang2025qwen3}, Mistral-7B \citep{jiang2023mistral7b} and InternLM2.5-7B
\citep{cai2024internlm2}. Significance uses question-paired McNemar tests
\citep{mcnemar1947note,dietterich1998approximate}. Appendix~\ref{app:settings} specifies
all of these exactly, including the three training-free references \emph{Random},
\emph{All-Tools} and \emph{Tools w/ Desc}.

\subsection{A1: \method{} outperforms sampled and offline baselines}
\label{sec:exp:main}

\definecolor{cBest}{HTML}{FFE0B2}
\definecolor{cSecond}{HTML}{E3F2FD}
\definecolor{cPos}{HTML}{2E7D32}
\definecolor{cNeg}{HTML}{C62828}
\newcommand{\gain}[1]{\cellcolor{cPos!#1}}
\begin{table}[t]
\centering
\caption{\textbf{Main comparison.} Accuracies (\%). Qwen3-8B supplied the training rewards; the
other four reasoners receive the same policy without adaptation. Trained rows share data,
prompts and adapters; GRPO keeps \citeauthor{tong2026autagent}'s differential reward while
\method{} optimizes Eq.~\ref{eq:reward} (\secref{sec:exp:reward} examines the parsimony term).
Per cell: best \textbf{bold}, second \underline{underlined}; ties share a mark.}
\label{tab:main}
\vspace{4pt}
\definecolor{cBest}{HTML}{FFF3E0}
{\scriptsize
\setlength{\tabcolsep}{2.6pt}
\renewcommand{\arraystretch}{1.18}
\resizebox{\linewidth}{!}{%
\begin{tabular}{llcccccccc}
\toprule
Benchmark & Reasoner & No-Tool & Random & All-Tools & Tools w/ Desc & SFT & DPO & GRPO & \cellcolor{cBest}\textbf{\method{}}\\
\midrule
\multirow{5}{*}{\rotatebox[origin=c]{90}{\textbf{GenomeQA}}}
 & Qwen3-8B\textsuperscript{$\dagger$} & 39.33 & 45.32 & \underline{49.44} & 47.77 & 42.17 & 39.33 & 45.43 & \cellcolor{cBest}\textbf{52.65}\\
 & Qwen2.5-1.5B & 38.25 & 40.78 & 42.92 & \textbf{44.07} & 40.72 & 38.33 & 42.87 & \cellcolor{cBest}\underline{43.79}\\
 & Qwen2.5-7B & 37.86 & 44.93 & \underline{48.91} & 47.10 & 40.25 & 38.22 & 46.74 & \cellcolor{cBest}\textbf{50.58}\\
 & Mistral-7B & 37.83 & 34.82 & 23.09 & \underline{41.00} & 38.08 & 37.69 & 35.68 & \cellcolor{cBest}\textbf{42.90}\\
 & InternLM2.5-7B & 38.36 & 45.38 & 47.94 & \underline{48.66} & 42.48 & 38.77 & 48.05 & \cellcolor{cBest}\textbf{52.53}\\
\midrule
\multirow{5}{*}{\rotatebox[origin=c]{90}{\textbf{GenBench-X}}}
 & Qwen3-8B\textsuperscript{$\dagger$} & 41.10 & 54.20 & \underline{59.00} & \underline{59.00} & 46.60 & 41.30 & 51.90 & \cellcolor{cBest}\textbf{65.20}\\
 & Qwen2.5-1.5B & 41.40 & 50.50 & \underline{58.50} & 54.20 & 42.00 & 41.40 & 48.80 & \cellcolor{cBest}\textbf{60.00}\\
 & Qwen2.5-7B & 42.10 & 57.30 & \underline{66.60} & 59.90 & 51.50 & 42.30 & 57.50 & \cellcolor{cBest}\textbf{68.90}\\
 & Mistral-7B & 40.30 & 43.50 & 45.40 & \underline{52.00} & 41.30 & 40.30 & 44.00 & \cellcolor{cBest}\textbf{58.20}\\
 & InternLM2.5-7B & 39.70 & 54.20 & \underline{63.10} & 59.70 & 50.00 & 39.90 & 54.10 & \cellcolor{cBest}\textbf{66.90}\\
\midrule
\multirow{5}{*}{\rotatebox[origin=c]{90}{\textbf{BM4}}}
 & Qwen3-8B\textsuperscript{$\dagger$} & 43.20 & 48.00 & \underline{53.30} & 47.30 & 43.50 & 43.10 & 52.30 & \cellcolor{cBest}\textbf{54.80}\\
 & Qwen2.5-1.5B & 42.60 & 47.00 & \underline{50.50} & 47.30 & 42.70 & 42.70 & 48.80 & \cellcolor{cBest}\textbf{51.80}\\
 & Qwen2.5-7B & 43.60 & 48.00 & \underline{55.00} & 47.50 & 44.30 & 43.50 & 52.90 & \cellcolor{cBest}\textbf{55.30}\\
 & Mistral-7B & 42.30 & 43.70 & 42.80 & \textbf{48.90} & 41.70 & 42.30 & \underline{45.20} & \cellcolor{cBest}\textbf{48.90}\\
 & InternLM2.5-7B & 42.60 & 48.90 & \underline{53.70} & 47.10 & 45.20 & 42.40 & 51.70 & \cellcolor{cBest}\textbf{54.70}\\
\bottomrule
\multicolumn{10}{l}{\textsuperscript{$\dagger$}Training reasoner; the other four columns transplant the same policy without adaptation.}
\end{tabular}}}
\end{table}

Table~\ref{tab:main} reports every selection strategy under matched conditions.
\bb{\method{} is the best or tied-best learned or deployable method in $14$ of the $15$
cells}, exceeding GRPO by $6.75$ points on average.
More importantly, \method{} outperforms GRPO in all $15$ benchmark--reasoner cells,
showing that the gain persists across frozen reasoners rather than being tied to the
Qwen3-8B reasoner that supplied the training rewards.
On GenomeQA \method{} also calls fewer tools than GRPO ($1.40$ vs.\ $2.36$ per question)
and uses $3.5$--$4.1\times$ fewer input tokens than exhaustive All-Tools
(Table~\ref{tab:economy}, Appendix~\ref{app:full}).

\bb{What the selections look like question by question.} On the $418$ questions where
\method{} is right and GRPO is wrong, the selected tool typically returns a per-option
\emph{contrast} rather than a decisive reading of the winning option alone
(Figure~\ref{fig:analysis}c; Appendix~\ref{app:full} works through three case studies).

\begin{figure}[t]
\centering
\includegraphics[width=\linewidth]{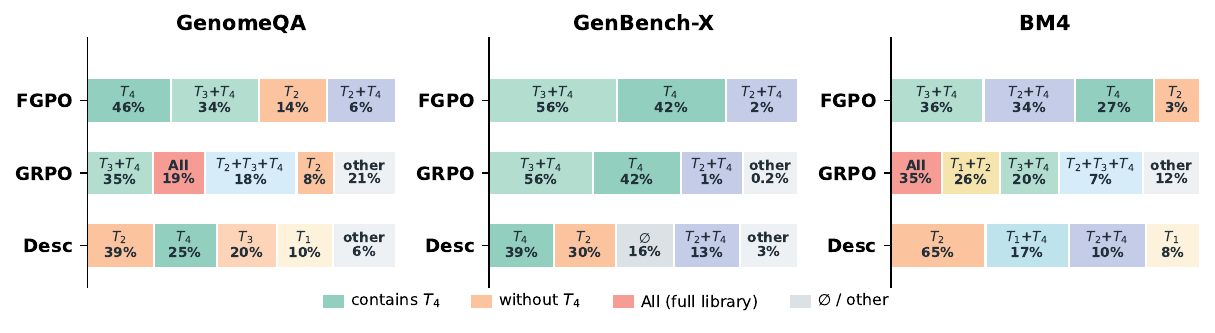}
\caption{\textbf{What each selector actually calls.} Top subset combinations per benchmark.
Teal/blue marks subsets containing $T_4$, orange those without, red the full library,
and gray the empty set or residual subsets. Widths are compressed for legibility; labels
give the displayed shares. \method{} concentrates on a few $T_4$-containing subsets,
whereas GRPO spreads broadly, including \emph{All} on $19\%$ of GenomeQA and $35\%$ of BM4.}
\label{fig:distribution}
\end{figure}

\subsection{A2: \method{} learns query-dependent genomic evidence routing}
\label{sec:exp:shapes}
\bb{They fail in three different shapes.} Read in aggregate
(Figure~\ref{fig:distribution}), on GenomeQA the prompted selector calls exactly one tool
on $93.6\%$ of questions, selecting a singleton subset on nearly every question, while GRPO
calls two or more on $81\%$ and spends three or four tools on $37.7\%$. \method{} does neither: it
never exceeds two tools, drops $T_1$ entirely ($0.0\%$ on all three benchmarks, against
GRPO's $21$--$64\%$) while concentrating on $T_4$, the tool that uniquely rescues
the most questions ($308$), calling it on $86.4\%$ of GenomeQA questions and at least
$96\%$ on the other two. Dropping $T_1$ is a decision, not a
free lunch: removing it from the library costs the per-question oracle $2.1$--$3.3$ points
across the three benchmarks (Table~\ref{tab:libablation}), so \method{} is giving up
reachable questions in exchange for never paying $T_1$'s cost on the rest. None of this was supervised (tool
slots are anonymous), so the shape of the policy is a statement about what the reward could
be made to reveal.

\bb{Offline objectives collapse onto the marginal mode.}
On the GenomeQA test set the empty subset is reward-optimal on $61.9\%$ of questions (the
reasoner is often right unaided, and the parsimony term then favors calling nothing).
SFT clones this marginal: its argmax is the empty set on $92.8\%$ of questions. DPO
collapses entirely: across three $\beta$ values and seven checkpoints it selects
$0.00$--$0.10$ tools and lands on the no-tool floor ($39.33$ vs.\ $39.28$), because
$58.1\%$ of chosen responses are the empty string and a pairwise ranking objective is
satisfied by the mode \citep{tong2026autagent}. \bb{GRPO does not collapse; it
starves}, which is Q3.

\begin{figure}[t]
\centering
\includegraphics[width=\linewidth]{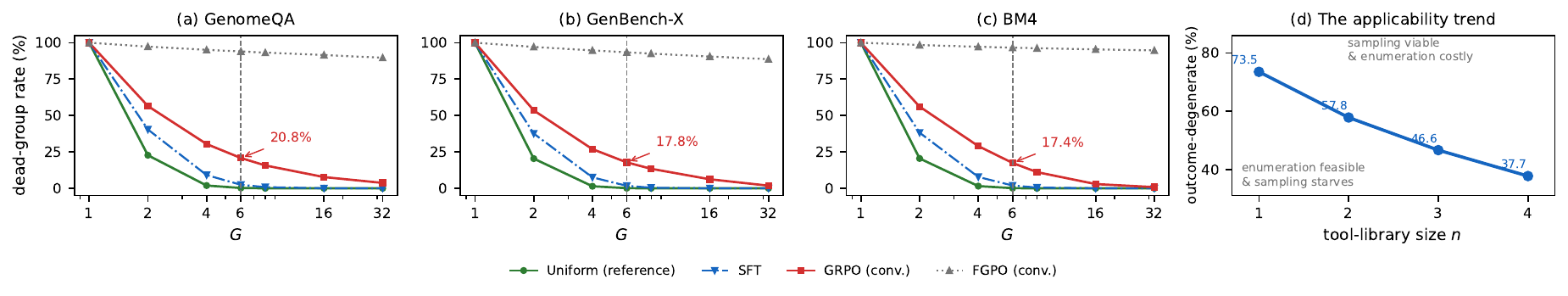}
\caption{\textbf{Dead-group rate replicates across benchmarks.} (a)--(c) Closed-form rate
vs.\ group size $G$; dashed line marks GRPO's $G{=}6$. Panel (a) adds the offline and exact
arms to Figure~\ref{fig:teaser}b; (b),(c) repeat it on the held-out benchmarks.
(d) \emph{Outcome-degenerate} fraction (all subsets induce the same correctness outcome)
vs.\ library size, averaged over sub-libraries.}
\label{fig:deadpanels}
\end{figure}

\begin{figure}[t]
\centering
\includegraphics[width=\linewidth]{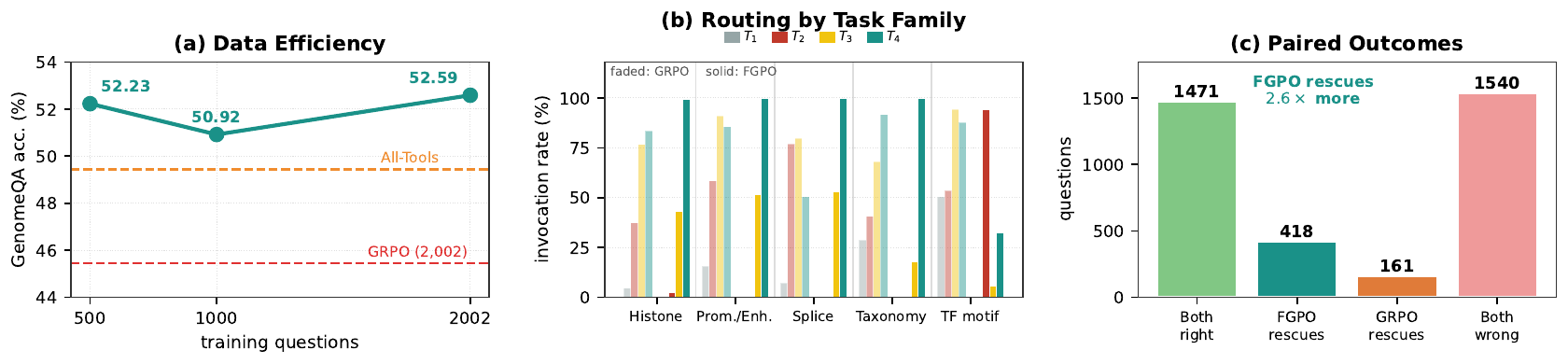}
\caption{\textbf{Additional analysis.} (a) $500$ training questions already beat GRPO
trained on $2{,}002$ questions and the training-free All-Tools baseline. (b) Per-tool invocation rate \emph{within} each task family
(Figure~\ref{fig:distribution} aggregates over them). GRPO calls every tool at a similar
rate whatever the question type; \method{} routes ($T_2$ for TF motif, $T_4$
elsewhere) and never calls $T_1$ on any benchmark against GRPO's $21$--$64\%$: a learned
refusal. (c) \method{} rescues $2.6\times$ as many questions as GRPO.}
\label{fig:analysis}
\end{figure}

\subsection{A3: Sampling starves \emph{because} training succeeds}
\label{sec:exp:mechanism}
For the closed-form diagnostic, let $p_\theta^{\mathrm{gen}}(a\,|\,s)$ denote the policy's
generation-induced distribution over the $16$ valid subset strings. The probability that
$G$ independent draws all land in one reward class, a dead group, is closed-form:
$P_{\text{dead}}(s) = \sum_{v}\big(\sum_{a: r(s,a)=v} p_\theta^{\mathrm{gen}}(a|s)\big)^{G}$. At GRPO's
$G{=}6$: $0.2\%$ of questions are dead under a uniform reference policy, $20.8\%$ under the
converged GRPO policy, $104\times$ the uniform-reference rate, replicated at
$17.8\%$/$17.4\%$
($178\times$/$174\times$) on GenBench-X/BM4 (Figure~\ref{fig:deadpanels}). These rates use
the reward classes of Eq.~\ref{eq:reward}, the finest any arm induces. Under the coarser
differential reward GRPO actually trains on, the dead fraction reaches $79.6\%$ at
convergence, so these figures are conservative.

\bb{The starvation is not benign.} A vanishing signal would be unremarkable if it dried up
only where the policy had already succeeded. It does not: dead-group probability is
empirically near-independent of correctness ($54\%$ of the dead mass sits on wrong answers),
and $6.7\%$ of all questions are simultaneously dead, wrong, and repairable by some subset
in the policy's own action space. The policy stops learning on them before it solves them,
while the exact estimator's signal on them never vanishes by sampling.

\begin{figure}[t]
\centering
\includegraphics[width=\linewidth]{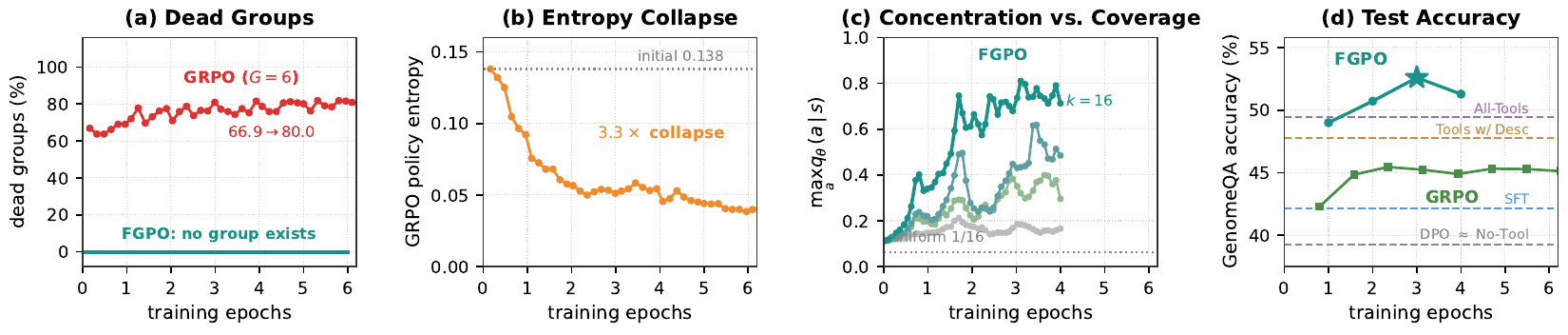}
\caption{\textbf{The mechanism, measured live during training.}
(a) The observed dead-group rate rises throughout GRPO training, while the exact
objective has no sampled groups by construction.
(b) GRPO's policy entropy decreases in parallel.
(c) \method{} also becomes highly concentrated while retaining complete action coverage.
(d) Test accuracy against the three training-free references in
Table~\ref{tab:main}: \method{} surpasses all three, whereas GRPO surpasses only
\emph{Random}.}
\label{fig:dynamics}
\end{figure}

Live telemetry under the coarser differential reward actually used by GRPO shows the same
dynamic during temperature-$1.0$ generation; malformed completions additionally join the
same reward class. Over the reported GRPO run the observed dead-group fraction rises from
$66.9\%$ to $80.0\%$ (Figure~\ref{fig:dynamics}a) as its policy entropy collapses by
$3.3\times$ (Figure~\ref{fig:dynamics}b). These observations are consistent with the mechanism formalized in Appendix~\ref{app:theory}: greater concentration of probability mass over reward classes increases the probability of a dead group. \method{} also concentrates, reaching $0.78$ on
its top action against a uniform $1/16$ (Figure~\ref{fig:dynamics}c), while avoiding sampled
dead groups. GRPO's test accuracy (Figure~\ref{fig:dynamics}d) peaks at $37.5\%$ of training and is
flat thereafter while its training reward gains a further $21\%$, so the objective keeps
improving after it has stopped buying accuracy, and it never reaches the two strongest
training-free references of Table~\ref{tab:main} (All-Tools, Tools w/ Desc), which
\method{} clears along with the rest.

The same tables give a library-size trend. Call a question \emph{outcome-degenerate} if all
$2^{n}$ subsets induce the same correctness outcome, so tool selection cannot alter whether
the frozen reasoner is right. Averaging over all $\binom{4}{n}$ sub-libraries of each
size, averaging over tool identity, Figure~\ref{fig:deadpanels}d shows this
fraction falling from $73.5\%$ at one tool to $37.7\%$ at four. Larger libraries thus
expose more outcome-diverse actions at the same time that exhaustive enumeration becomes
more expensive. The two estimators' applicability regions are complementary: this paper
characterizes the regime where exact optimization is feasible and sampled optimization is
most information-starved.

\subsection{A4: A controlled coverage intervention: the \texorpdfstring{$k$}{k}-subset dose--response}
\label{sec:exp:kcurve}

\begin{wraptable}{r}{0.505\linewidth}
\vspace{-11pt}
\caption{\textbf{The coverage dial.} Accuracy by epoch when the objective sees only $k$
uniformly drawn actions per visit. Everything else held fixed.}
\label{tab:kcurve}
\vspace{2pt}
{\footnotesize
\setlength{\tabcolsep}{2.8pt}
\begin{tabular*}{\linewidth}{@{\extracolsep{\fill}}lccccc}
\toprule
$k$ & ep1 & ep2 & ep3 & ep4 & peak\\
\midrule
$2$ & 47.55 & 49.14 & 49.58 & 49.44 & 49.58\\
$4$ & 47.27 & 50.25 & 50.47 & 45.24 & 50.47\\
$8$ & 49.14 & 50.89 & 51.39 & 50.56 & 51.39\\
$16$ (\method{}) & 49.00 & 50.72 & \best{52.59} & 51.28 & \best{52.59}\\
\midrule
GRPO ($G{=}6$) & \multicolumn{4}{c}{---} & 45.46\\
\bottomrule
\end{tabular*}}
\vspace{-8pt}
\end{wraptable}

The dead-group analysis is observational; we now test it with an intervention. We hold
the data, ordering, hyperparameters and action-scoring construction fixed and vary only the
number of candidate actions the estimator is shown: at each visit it sees a uniformly drawn
subset of $k$ of them, with $q_\theta$ renormalized on that subset, so each $k$ is its own
objective and $k{=}16$ recovers \method{}. Policy-model compute per step is unchanged,
since all candidate strings are scored in every arm. Accuracy at the (dev-selected)
peak epoch is strictly monotone in $k$: $49.58 \to 50.47 \to 51.39 \to 52.59$ for
$k = 2, 4, 8, 16$ (paired McNemar: $k{=}16$ over $k{=}2$, $p < 10^{-4}$; over $k{=}8$,
$p = 0.011$; Figure~\ref{fig:teaser}c and Table~\ref{tab:kcurve}). Two readings
follow. First, what the estimator sees of the action space is what the method learns, the
controlled counterpart of \secref{sec:exp:mechanism}. Second, even $k{=}2$ ($49.58$) exceeds
six-rollout GRPO ($45.46$): uniform draws do not concentrate with the policy, consistent
with GRPO's deficit having two components, coverage \emph{and} sampling from a concentrated
distribution, though only coverage is varied here. The $k$ dial also spans a coverage--budget continuum: using only two frozen-reasoner reward evaluations per visit, one third of GRPO's, the
budgeted variant still leads by $4$ points.

\subsection{A5: Effect of parsimony on accuracy and tool usage}
\label{sec:exp:reward}

\begin{wraptable}{r}{0.44\linewidth}
\vspace{-11pt}
\caption{\textbf{Reward design.} The exact objective trained on Eq.~\ref{eq:reward} (correctness
plus the parsimony term) against correctness alone ($\lambda{=}0$), both scored over the
enumerated candidates on the $1{,}200$-question quick split. Independent runs with different initialization.}
\label{tab:reward}
\vspace{2pt}
{\footnotesize
\setlength{\tabcolsep}{3pt}
\resizebox{\linewidth}{!}{%
\begin{tabular}{@{}lcccc@{}}
\toprule
 & ep1 & ep2 & ep3 & ep4\\
\midrule
With parsimony, acc. & 47.25 & 48.33 & \best{51.17} & 49.75\\
Correctness only, acc. & 47.58 & 48.33 & 48.75 & 49.17\\
\midrule
With parsimony, tools & 1.04 & 1.53 & 1.61 & 1.19\\
Correctness only, tools & 2.04 & 2.33 & 2.06 & 2.19\\
\bottomrule
\end{tabular}}}
\vspace{-8pt}
\end{wraptable}

\bb{Penalizing redundancy primarily buys tool economy.} Table~\ref{tab:reward} retrains
the exact objective on correctness alone ($\lambda{=}0$ in Eq.~\ref{eq:reward}) against
the reported reward, which adds the parsimony term. The $\lambda{=}0$ arm is also the
closest exact-objective counterpart of GRPO's reward: on the enumerated candidates
\citeauthor{tong2026autagent}'s differential reward is half the $\pm1$ correctness reward
plus a per-question constant, to which the expected-reward term of Eq.~\ref{eq:exact} is
invariant, so that term has a proportional gradient under either; the entropy weight and
GRPO's sampling and group normalization lie outside this identity. Between the two arms accuracy differs modestly ($49.17$ vs.\ $51.17$ at
peak) but economy does not: without the term the policy invokes more tools at every
epoch ($2.0$--$2.3$ against $1.0$--$1.6$; $2.19$ vs.\ $1.19$ at epoch~4), as expected when
a right answer obtained with four tools scores exactly as well as the same answer with one.

\section{Related Work}
\label{sec:related}

\textbf{Tool-augmented genomic reasoning.}
Genomic foundation models provide specialized representations for nucleotide sequences
\citep{ji2021dnabert,zhou2024dnabert,dalla2025nucleotide,nguyen2023hyenadna,schiff2024caduceus,nguyen2024sequence},
and general-purpose LLMs benefit from access to such tools \citep{jin2024genegpt}; public
benchmarks for genomic tasks exist
\citep{zhou2024dnabert,grevsova2023genomic,de2022deepstarr}. These directions address
complementary parts of the problem: genomic models supply sequence-level computation, while
language models supply semantic reasoning. Our focus is the interface, deciding which
specialized genomic computations to expose for each question.

\textbf{RL-based tool selection.}
Teaching LLMs to call external tools spans self-supervised call insertion
\citep{schick2023toolformer}, prompted acting \citep{yao2022react,lu2023chameleon},
API-scale instruction tuning
\citep{qin2024toolllm,patil2024gorilla,shen2023hugginggpt}, and benchmarks of call
correctness \citep{li2023api}. Closest to us is the line that trains a
\emph{selection} policy for a frozen reasoner with RL: VisTA \citep{huang2025visualtoolagent}, AuTAgent
\citep{tong2026autagent}, and reward-shaped variants
\citep{qian2026toolrl,jin2025search}, all of which modify the reward or interaction
protocol while retaining sampled policy optimization. We adopt their frozen-reasoner setting but study the combinatorial subset-selection
regime: what happens when the complete tool-subset space is small enough to evaluate but
the optimizer continues to sample it? Our results show that the issue is not only estimator variance; policy concentration
progressively increases the probability of zero-advantage groups. Integrating over actions rather than sampling them has been explored for variance reduction
\citep{ciosek2018expectedpolicygradients,ciosek2020expected,asadi2017mean,kool2019buy,williams1992simple,ahmadian2024back}.
GRPO \citep{shao2024deepseekmath} and its successors
\citep{yu2026dapo,liu2025understanding} normalize within sampled groups; the resulting
zero-advantage behavior is noted as an efficiency concern \citep{yu2026dapo}. We
characterize its interaction with policy concentration on enumerable spaces, where it
becomes a structural mismatch rather than a mere inefficiency, and propose \method{} as a
practical solution for this regime: regime identification, failure mechanism, and LLM
instantiation.

\section{Conclusion and Future Work}
\label{sec:conclusion}
Genomic tools provide sequence-level evidence that language models cannot reliably
compute, but their usefulness is strongly query-dependent. We show that sampled group
optimization becomes increasingly information-starved as the policy concentrates over a
small enumerable tool space, causing reward collisions and vanishing advantages.
\method{} removes this sampling mismatch by optimizing the exact action expectation,
consistently improving reasoning across three genomic benchmarks and five frozen reasoners
while invoking fewer tools. While compact tool libraries suffice in the specialist regime
studied here, future work could extend \method{} to domains requiring larger libraries by
adaptively constructing compact, coverage-preserving candidate sets and retaining exact
optimization within each selected set.

\subsubsection*{AI use statement}
Generative AI tools were used to assist with writing: polishing prose for clarity and
concision. They were not used to
generate research ideas, design experiments, or draw conclusions. We have reviewed all
AI-assisted text and take responsibility for the final content of this work.

\subsubsection*{Ethics statement}
This work involves no human subjects and no clinical or individually identifiable genomic
data. All sequences come from public reference-organism corpora; the two held-out evaluation
suites are derivatives of those corpora and redistribution should respect each upstream
licence. The method learns only which standard sequence-analysis tools to invoke and neither
designs sequences nor adds capabilities beyond those the tools already provide.

\subsubsection*{Reproducibility statement}
All training and evaluation code, the exhaustive reward tables, per-question answer files
for every table row, and the failure log (including models that could not be evaluated on
our hardware, with root causes) will be released. Baselines use official TRL
implementations \citep{vonwerra2020trl}; every number in the paper regenerates from the
released tables by scripted lookup.

\bibliography{0908}
\bibliographystyle{iclr2027_conference}

\clearpage
\appendix
\begin{center}{\Large\textsc{Contents of Appendix}}\end{center}
\vspace{4pt}
\startcontents[appendix]
{\hypersetup{linkcolor=black}\printcontents[appendix]{}{1}{}}
\vspace{6pt}

\section{Algorithm}
\label{app:algorithm}
Algorithm~\ref{alg:fgpo} states \method{} end to end. Stage~1 is the one-off cost
(\secref{sec:method:table}); Stage~2 touches the frozen reasoner not at all, since every
reward it needs is a table lookup. The inner loop scores all $2^{n}$ candidate strings in
a single batched forward pass, requiring no policy-side sampling.

\begin{algorithm}[h]
\SetAlgoLined
\DontPrintSemicolon
\KwIn{questions $\mathcal{D}$; tool library $\{T_1,\dots,T_n\}$; frozen reasoner $R$;
LoRA-parameterized policy $q_\theta$; entropy weight $\beta$; parsimony weight $\lambda$}
\KwOut{trained selector $q_\theta$}
\tcp{Stage 1: exhaustive reward table (once)}
\ForEach{$s \in \mathcal{D}$}{
  \ForEach{$a \in \mathcal{A}=2^{\{T_1,\dots,T_n\}}$}{
    run the tools in $a$ on $s$'s sequence; query $R$ with the rendered evidence\;
    store $r(s,a)$ by Eq.~\ref{eq:reward}\;
  }
}
\tcp{Stage 2: exact policy optimization}
\For{epoch $=1,\dots,E$}{
  \ForEach{minibatch $B\subseteq\mathcal{D}$}{
    \ForEach{$s\in B$}{
      score every candidate string: $\ell_\theta(a\,|\,s)$ for all $a\in\mathcal{A}$
      (Eq.~\ref{eq:score})\tcp*{one batched forward pass}
      $q_\theta(\cdot\,|\,s)\leftarrow\mathrm{softmax}_{a\in\mathcal{A}}\,\ell_\theta(a\,|\,s)$\;
    }
    $\mathcal{L}(\theta)\leftarrow-\frac{1}{|B|}\sum_{s\in B}\Big[\sum_{a\in\mathcal{A}}
      q_\theta(a|s)\,r(s,a)+\beta\,H\big(q_\theta(\cdot|s)\big)\Big]$
      \tcp*{Eq.~\ref{eq:exact},~\ref{eq:entropy}}
    $\theta\leftarrow\theta-\eta\,\nabla_\theta\mathcal{L}(\theta)$
      \tcp*{every $a$ contributes; no group, no dead groups}
  }
  select the checkpoint on the held-out development split\;
}
\caption{\method{} (\methodlong{})}
\label{alg:fgpo}
\end{algorithm}

\section{Experiment Settings}
\label{app:settings}

\subsection{Prompts and action encoding}
The policy prompt lists $n$ anonymous tool slots (``\texttt{0: type1 (A)}'' style) and the
question; the answer is an indexed subset in \texttt{<answer></answer>} tags. Tool names,
descriptions, and any ``useful-for'' hints are withheld from the policy in every trained
arm (\method{}, GRPO, SFT, DPO); the Tools-w/-Desc baseline receives real names and
functional descriptions. Reasoner prompts concatenate the question, options, and the
rendered evidence of the selected subset; evidence rendering is shared verbatim across
every method and every table row.
\bb{Anonymous slots are not a handicap.} To check that the policy learns routing from
reward rather than from tool names, we retrain \method{} with the real tool names and
functional descriptions in its selection prompt. On a fixed $1{,}200$-question subsample of the
GenomeQA test set the two are nearly identical: $51.67$ with descriptions against
$51.75$ without, both at format rate ${\approx}1.0$. The anonymous encoding used
throughout therefore costs nothing, and the routing reported in \secref{sec:exp:shapes}
cannot have been read off the prompt.

\subsection{Sensitivity checks}
\label{app:sensitivity}
Table~\ref{tab:sensitivity} varies three choices one at a time on a fixed
$1{,}200$-question subsample of the GenomeQA test set, disjoint from the $492$-question
development split used for checkpoint selection. The exact objective changes little
under the tested entropy coefficient, since raising $\beta$ from $0.03$ to $0.08$ changes the
diagnostic peak by only $0.25$ points, and under a sharper listwise softmax,
$q_\tau(a\,|\,s)\propto\exp\!\big(\ell_\theta(a\,|\,s)/\tau\big)$ with $\tau{=}0.7$ in place of
Eq.~\ref{eq:score}. A $1.5$B selector in place of the
$7$B one still reaches $50.42$ against six-rollout GRPO's $44.25$, so the advantage is not
specific to the $7$B selector.

\begin{table}[h]
\centering
\caption{\textbf{Sensitivity of the exact objective.} Accuracy (\%) per epoch on the
$1{,}200$-question subsample of the GenomeQA test set; \emph{peak} is the maximum observed
accuracy on this diagnostic subsample and is reported only for sensitivity analysis, never
for checkpoint selection. Every row changes one factor from the reported configuration (first row).
GRPO scores $44.25$ here; a dash marks a cell not evaluated on this subsample.}
\label{tab:sensitivity}
\vspace{2pt}
{\footnotesize
\begin{tabular*}{\linewidth}{@{\extracolsep{\fill}}lccccc}
\toprule
Variant & ep1 & ep2 & ep3 & ep4 & peak\\
\midrule
Reported ($\beta{=}0.03$, $7$B policy) & 47.00 & 48.83 & \best{51.75} & --- & \best{51.75}\\
\midrule
Entropy $\beta = 0.08$ & 48.17 & 50.50 & \best{52.00} & 50.67 & 52.00\\
Listwise softmax temperature $0.7$ & 46.00 & 48.75 & \best{50.92} & 50.58 & 50.92\\
$1.5$B policy (same objective) & 47.33 & \best{50.42} & 48.42 & 49.58 & 50.42\\
\bottomrule
\end{tabular*}}
\end{table}

\subsection{A five-tool library: the policy rejects a harmful capability}
\label{app:fivetool}
To probe what happens when the library grows, we add a fifth tool $T_5$, a homology-search
module, while keeping $T_1$--$T_4$ in place, and retrain the exact objective over the
resulting $2^{5}=32$ subsets. $T_5$ searches the query against a local database of
labelled sequences by $k$-mer seeding followed by Smith--Waterman alignment
\citep{cock2009biopython}, reporting the closest matches with label, percent identity and
query coverage; the database is built from training and development sequences only, and
self-hits and near-identical matches ($>\!99.5\%$ identity) are dropped so the tool
cannot retrieve the query and hand back its own label. This tool is
a poor fit for the benchmark: used alone it scores $34.07$, below the $39.33$ no-tool floor,
so a selector that simply invokes everything available should be harmed by it. On the
full $3{,}590$-question test set the five-tool policy reaches $45.38$ at its final
checkpoint, above exhaustive invocation of all five tools ($44.51$) and above the best
post-hoc fixed subset ($45.18$); accuracy rises monotonically over the four epochs
($43.04\to44.15\to44.51\to45.38$), so the final checkpoint is also the best observed one; no checkpoint is selected
using test performance. The per-tool invocation rates explain why: the policy calls the motif
scanner on $61.9\%$ of questions and the genomic encoder on $14.3\%$, but the harmful new
module on only $2.8\%$, and composition statistics, the weakest of the original four, on
$0.0\%$. Learning from complete action-level feedback
therefore includes learning which capabilities to \emph{refuse}, not only which to prefer.
These runs use a separately constructed reward table for the five-tool action space, and
every reference point quoted above ($34.07$, $39.33$, $44.51$, $45.18$) is read from that
same $32$-mask table. That table was built with an earlier version of the genomic encoder
$T_4$: on the same $3{,}590$ questions, the subsets without $T_4$ agree with the four-tool
table to within $0.1$ points, while every subset containing it scores lower ($T_4$ alone
$40.25$ vs.\ $47.77$). All comparisons are therefore internal to this
table, and the runs are intended as a library-expansion stress test rather than a matched
performance comparison with the four-tool setting of Table~\ref{tab:main}.

\subsection{Hyperparameters}
Table~\ref{tab:hyperparam} lists every value; \secref{app:sensitivity} reports what happens
when they are varied. Two details do not fit the table. The reported GRPO run is the configuration that
faithfully reproduces \citeauthor{tong2026autagent}'s published setting, fixed \emph{a
priori} rather than selected by score; four further independent configurations (varying
$G$, temperature and data mix) span $43.90$--$45.57$. And checkpoint selection
for the four-tool trained methods of the main comparison (\method{}, GRPO, SFT) uses the
same $492$-question development split, with test accuracy reported at the dev-selected
checkpoint; for \method{}, dev and test agree on the peak (ep3: $52.24$ dev / $52.59$
test). DPO is the exception: every $\beta$ and every checkpoint collapses to the no-tool
floor, so we report the best cell of its entire sweep, which is more generous than dev
selection and still leaves it at $39.33$. The five-tool run of \secref{app:fivetool} has no
development sweep of its own and is reported at its final epoch.

\label{app:hyperparam}
The trained arms share the same backbone, LoRA configuration, training data, prompt
format, and data order; objective-specific optimization settings and training durations
are listed explicitly below.

\begin{table}[h]
\centering
\caption{Key hyperparameters.}
\label{tab:hyperparam}
\vspace{2pt}
{\small
\resizebox{\linewidth}{!}{%
\begin{tabular}{llc}
\toprule
Hyperparameter & Value & Shared across arms?\\
\midrule
Base model & Qwen2.5-7B-Instruct & \checkmark\\
LoRA rank / alpha & 16 / 32 & \checkmark\\
Learning rate & $10^{-5}$ (cosine) & \checkmark\\
\method{}/SFT/DPO batch size & 16 prompts & \method{}/SFT/DPO\\
GRPO prompts per step & 64 ($\times 6$ rollouts) & GRPO only\\
\method{}/SFT/DPO training & 4 epochs & \method{}/SFT/DPO\\
GRPO training & 200 steps ($\approx 6.26$ epochs as logged by the trainer) & GRPO only\\
Gradient checkpointing & Yes & \checkmark\\
Entropy bonus $\beta$ (\method{}) & 0.03 & ---\\
Parsimony $\lambda$ (Eq.~\ref{eq:reward}) & 0.10 & \method{}/SFT/DPO\\
GRPO rollouts $G$ & 6 & GRPO only\\
GRPO temperature & 1.0 & GRPO only\\
DPO $\beta$ & 0.3 (all of 0.1/0.3/0.5 collapse) & DPO only\\
Controlled candidate-scoring eval.\ & per-token mean (Eq.~\ref{eq:score}) & all trained arms\\
Dev-set checkpoint selection & 492 questions & \method{}/GRPO/SFT\\
\bottomrule
\end{tabular}}}
\end{table}

\subsection{Anonymous slot encoding examples}
\label{app:slot_examples}
To verify that the policy cannot read tool identity from the prompt, we show the full
input-output pair for one training question.

\paragraph{Policy input.}
\begin{quote}\ttfamily\small
[System] You are an expert agent specialized in selecting tools to solve genomic
reasoning tasks. You are provided with access to 4 tools, indexed from 0 to 3.\\[4pt]
[User] Available tools:\\
0: type1 (A)\\
1: type2 (B)\\
2: type3 (C)\\
3: type4 (D)\\[4pt]
I have extracted a Human DNA sequence\ldots Which of the following best describes it?\\
A: enhancer region \quad B: promoter region \quad C: splice site \quad D: coding region\\[4pt]
Select the index number(s)\ldots enclosed in <answer></answer> tags.
\end{quote}

\paragraph{Policy output (FGPO, converged).}
\begin{quote}\ttfamily\small
<answer>3</answer>
\end{quote}
Tool 3 maps to $T_4$ (genomic expert). The policy has no way to know this from the
prompt: it learned the mapping entirely from reward.

\paragraph{Policy output (GRPO, converged).}
\begin{quote}\ttfamily\small
<answer>0,1,2,3</answer>
\end{quote}
The converged GRPO policy selects all four tools on $19\%$ of GenomeQA questions
(Figure~\ref{fig:distribution}).

\subsection{Baselines}
\emph{Random} draws one subset uniformly from all $2^{n}$, including the empty one, per
question at a fixed seed. \emph{All-Tools} always invokes the full library.
\emph{Tools w/ Desc} is the same unadapted Qwen2.5-7B-Instruct backbone used as the
policy, given the real tool names and functional descriptions in its selection prompt.
GRPO is evaluated at checkpoints $\{25,\dots,200\}$ with the reported checkpoint
selected on dev; DPO at the best of a $\beta\in\{0.1,0.3,0.5\}$ sweep over seven
checkpoints, taking the best cell outright rather than a dev-selected one
(Appendix~\ref{app:fairness}).

\subsection{Reward}
Eq.~\ref{eq:reward}: $\pm1$ on the correctness of the reasoner's answer under the subset,
plus $\lambda(1-2|a|/n)$ with $\lambda=0.10$, so the reward runs in $[-1.1,+1.1]$ and
parsimony can never outrank being right. The GRPO arm instead uses the differential reward of \citet{tong2026autagent},
\begin{equation*}
r_{\mathrm{diff}}(s,a)=\mathbf{1}\!\left[\hat{y}(s,a)=y^{\star}\right]
-\mathbf{1}\!\left[\hat{y}(s,\emptyset)=y^{\star}\right]\in\{-1,0,+1\},
\end{equation*}
i.e.\ $+1$ when tools rescue a wrong no-tool answer, $-1$ when they break a right one and
$0$ otherwise, combined as $0.95\,r_{\mathrm{diff}}+0.05\,r_{\mathrm{fmt}}$ with
$r_{\mathrm{fmt}}\in\{0,1\}$ marking a parseable completion, faithful to that paper. Every
one of the $2^{n}$ enumerated candidates is parseable by construction, so
$r_{\mathrm{fmt}}$ is constant over them and does not change the reward partition used in
the closed-form dead-group analysis. The binary arm of Table~\ref{tab:reward} is
Eq.~\ref{eq:reward} with $\lambda=0$: $\pm1$ on correctness, no parsimony term. Every arm reuses the same precomputed frozen-reasoner outcomes for all subsets of every
training question (\secref{sec:method:table}); \method{}, SFT and DPO derive
Eq.~\ref{eq:reward} from them, GRPO derives its differential reward.

\subsection{Benchmark construction and split hygiene}
For \emph{GenomeQA}, evaluation uses the promoter/enhancer, splice-site, taxonomy,
histone-mark and TF-motif task families. \emph{Training questions} are built separately from the Nucleotide
Transformer downstream tasks (enhancers, promoter\_all, splice\_sites\_all, four histone
marks) in the same question shapes GenomeQA uses, so taxonomy and TF-motif families are
first seen at test time. \emph{GenBench-X} draws $200$ questions from each of five Genomic
Benchmarks subsets: \texttt{demo\_human\_or\_worm}, \texttt{human\_ensembl\_regulatory},
\texttt{human\_ocr\_ensembl}, \texttt{drosophila\_enhancers\_stark},
\texttt{demo\_coding\_vs\_intergenomic}, two of them non-human. \emph{BM4} draws $250$
from each of four sources: bacterial $\sigma^{70}$ promoters, a bacteria-vs-archaea
taxonomy set, DeepSTARR fly enhancers, and mouse Ensembl enhancers. Integer labels carry
no semantics in these corpora, so every label mapping was established by measurement (GC
content, in-frame stop-codon depletion, CpG observed/expected, cross-subset sequence
identity) rather than from the dataset name; two turn out to be the reverse of what the
name suggests. Sequences beyond $1{,}000$\,bp are centre-cropped to GenomeQA's ceiling.
Sequence-level disjointness between the training pool and each evaluation suite is checked
before any file is written, and each task slot's kNN reference set is rebuilt from that
source's own training split.

\subsection{Tool implementations}
$T_1$ \emph{sequence composition}: GC content, $k$-mer and codon statistics via
Biopython/NumPy; nothing trained. $T_2$ \emph{motif scanning}: JASPAR matrices, scanned
with FIMO from the MEME suite where available and otherwise with a Biopython PSSM scan
over matrices from \texttt{pyjaspar}. $T_3$ \emph{splice-site analysis}: GT/AG only
enumerates candidate positions; MaxEntScan scores them ($9$-mer donor, $23$-mer acceptor
models). $T_4$ \emph{genomic expert}: mean-pooled embeddings from a frozen
\texttt{nucleotide-transformer-v2-50m-multi-species} encoder
\citep{dalla2025nucleotide} with a cosine-metric $k$NN ($k{=}15$) over a labelled
reference set built from the corresponding task's \emph{training} split only, with neither
test sequences nor test labels included in the reference set, and the encoder is never
fine-tuned. Every
tool renders into the same structured evidence block for every method and table row.

\subsection{Infrastructure and models that could not be evaluated}
All experiments ran on Ascend 910B NPUs with vLLM-Ascend serving \citep{kwon2023efficient}. For
reproducibility we record models excluded for stack reasons: GLM-4-9B (fused-RoPE kernel
lacks partial-rotary support), Gemma-2-9B (\texttt{rope\_theta} config incompatibility in
the serving stack), Yi-1.5-9B (weights unavailable). None were excluded for score reasons.

\section{Full Transfer Results}
\label{app:full}

\subsection{Case study}
Table~\ref{tab:case} shows three representative questions where the frozen reasoner fails
without tools but is rescued by \method{}'s tool selection.

\begin{table}[h]
\centering
\caption{\textbf{Case study.} Three questions the reasoner gets wrong, rescued by
\method{}'s tool call. Questions and tool outputs are quoted from the exhaustive reward
table; interpretations are summarized for clarity.}
\label{tab:case}
\vspace{3pt}
{\footnotesize
\setlength{\tabcolsep}{3.5pt}
\renewcommand{\arraystretch}{1.12}
\resizebox{\linewidth}{!}{%
\begin{tabular}{@{}p{0.22\linewidth}|p{0.38\linewidth}|p{0.32\linewidth}@{}}
\toprule
\multicolumn{1}{c|}{\textbf{Qwen3-8B (No-Tool)}} & \multicolumn{2}{c}{\textbf{\method{} (Ours)}}\\
\cmidrule(r){1-1}\cmidrule(l){2-3}
\multicolumn{1}{c|}{\textbf{Wrong Answer}} & \multicolumn{1}{c|}{\textbf{Tool Call}} &
\multicolumn{1}{c}{\textbf{Final Answer}}\\
\midrule
\rowcolor{grouprow}[0pt][0pt]\multicolumn{3}{@{}p{0.985\linewidth}@{}}{\textbf{Ex1 (TF motif):}
``Which Human DNA sequence is a target for SRF?''
\textcolor{gtgreen}{[GT: D]}}\\
Picks \textcolor{wrongred}{A}; nothing separates the candidates. &
\texttt{motif\_scanner:} \{A: JUN, FOS, TEAD4 | B: STAT1 | C: SP1
| \textbf{D: SRF}\} &
\textcolor{gtgreen}{D} carries the named motif.\\
\midrule
\rowcolor{grouprow}[0pt][0pt]\multicolumn{3}{@{}p{0.985\linewidth}@{}}{\textbf{Ex2 (Taxonomy):}
``Select the DNA sequence derived from a Virus genome.''
\textcolor{gtgreen}{[GT: B]}}\\
Picks \textcolor{wrongred}{A}, the most eukaryotic. &
\texttt{genomic\_expert:} \{A: Euk .82 | \textbf{B: Virus .37}
| C: Euk .53 | D: Euk .84\} &
\textcolor{gtgreen}{B} is the only non-eukaryote.\\
\midrule
\rowcolor{grouprow}[0pt][0pt]\multicolumn{3}{@{}p{0.985\linewidth}@{}}{\textbf{Ex3 (Splice site):}
``Which Human DNA sequence contains a functional Only Acceptor?''
\textcolor{gtgreen}{[GT: B]}}\\
Picks \textcolor{wrongred}{A}, which has no splice site at all. &
\texttt{genomic\_expert:} \{A: none | \textbf{B: Only Acceptor .53}
| C: none | D: none\} &
\textcolor{gtgreen}{B} is the only candidate with a site.\\
\bottomrule
\end{tabular}}}
\end{table}

\begin{table}[h]
\centering
\caption{\textbf{Accuracy and cost.} Upper: all methods on GenomeQA under the training reasoner.
Lower: \method{} against exhaustive invocation on all three benchmarks. Input tokens count
question $+$ rendered evidence, measured for \method{} and All-Tools.}
\label{tab:economy}
\vspace{2pt}
{\footnotesize
\setlength{\tabcolsep}{4pt}
\begin{tabular*}{\linewidth}{@{\extracolsep{\fill}}lccccc}
\toprule
GenomeQA & No-Tool & GRPO & Tools w/ Desc & All-Tools & \textbf{\method{} (ours)}\\
\midrule
Accuracy & 39.28 & 45.46 & 47.77 & 49.39 & \best{52.62}\\
Tools / question & 0.00 & 2.36 & 1.03 & 4.00 & 1.40\\
Input tokens & -- & -- & -- & 4{,}486 & 1{,}300\\
\bottomrule
\end{tabular*}}

\vspace{5pt}
{\footnotesize
\setlength{\tabcolsep}{4pt}
\begin{tabular*}{\linewidth}{@{\extracolsep{\fill}}lccc}
\toprule
\method{} vs.\ All-Tools & GenomeQA & GenBench-X & BM4\\
\midrule
\method{} tool calls & 1.40 & 1.58 & 1.70\\
Call ratio & 2.86$\times$ & 2.54$\times$ & 2.36$\times$\\
Input-token ratio & 3.45$\times$ & 4.14$\times$ & 3.65$\times$\\
\bottomrule
\end{tabular*}}
\end{table}

\subsection{Where the gain comes from}
\label{app:bytask}
Table~\ref{tab:bytask} splits the GenomeQA column of Table~\ref{tab:main} into its five
task families. Two things are visible only at this resolution. First, the gain is not
uniform: \method{} adds $20.6$ points over the unaided reasoner on taxonomy and $23.0$ on
TF motif, but only $3.5$ on splice sites and $3.5$ on histone marks, the families
where the per-question oracle itself is lowest ($68.7$ and $67.7$), so the library simply
carries less signal there. Second, the average number of tools the policy calls
\emph{rises} as the family gets harder, from $1.18$ on taxonomy to $1.53$ on splice sites:
the policy spends its budget where a single tool does not settle the question. Nothing in
the reward asks for this (Eq.~\ref{eq:reward} penalises tools uniformly), so it is a
property the exact objective discovers rather than one it is told.

\begin{table}[h]
\centering
\caption{\textbf{GenomeQA by task family} ($n{=}718$ each, Qwen3-8B reasoner). Accuracies
(\%); \emph{tools} is \method{}'s mean subset size on that family.}
\label{tab:bytask}
\vspace{2pt}
{\footnotesize
\begin{tabular*}{\linewidth}{@{\extracolsep{\fill}}lccccc}
\toprule
Family & No-Tool & Tools w/ Desc & GRPO & \method{} & Oracle\\
\midrule
Taxonomy          & 46.24 & 55.15 & 56.55 & \best{66.85} & 81.6\\
TF motif           & 38.58 & 61.42 & 46.80 & \best{61.56} & 85.8\\
Promoter/enhancer & 36.21 & 35.65 & 43.31 & \best{52.09} & 83.3\\
Splice site       & 38.02 & \best{44.71} & 40.53 & 41.50 & 68.7\\
Histone mark      & 37.60 & \best{41.78} & 40.11 & 41.09 & 67.7\\
\midrule
\multicolumn{6}{l}{\emph{\method{} mean tools:} taxonomy $1.18$, TF motif $1.32$, prom./enh.\ $1.51$, histone $1.45$, splice $1.53$}\\
\bottomrule
\end{tabular*}}
\end{table}

\subsection{The transfer grid, read as shapes}
\label{sec:exp:transfer}
\begin{figure}[h]
\centering
\includegraphics[width=0.94\linewidth]{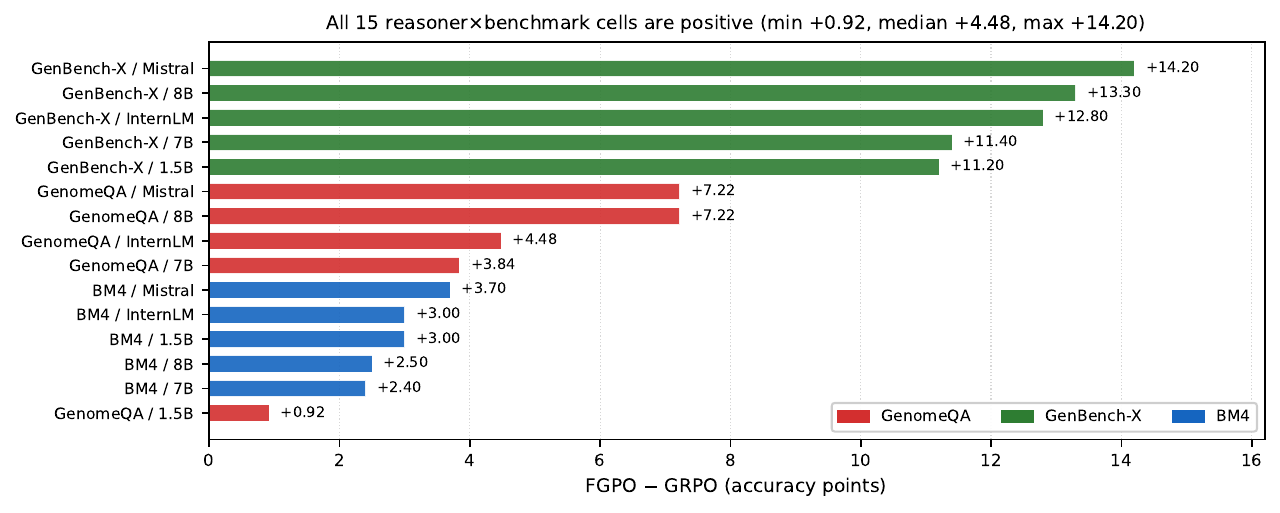}
\caption{\method{} $-$ GRPO for all $15$ reasoner$\times$benchmark cells, sorted. No cell
is negative. Colour denotes benchmark.}
\label{fig:deltacells}
\end{figure}

The four transfer columns of Table~\ref{tab:main} were produced by transplanting the
policy trained against Qwen3-8B rewards into a different frozen reasoner, with no
adaptation, no retuning and no access to the new reasoner during training. That this
works at all is not obvious, since a selection policy could easily have learned which subsets
suit one particular reader, so the result worth stating is the margin's consistency
rather than its size: \method{}$-$GRPO is positive in all $15$
reasoner$\times$benchmark cells (min $+0.92$, median $+4.48$, max $+14.20$;
Figure~\ref{fig:deltacells}), and no reasoner and no benchmark supplies a counterexample.
DPO tracks the no-tool floor within $0.41$ points in all $15$ cells, consistent with its
empirical collapse onto the empty subset, which turns that collapse into a
five-reasoner-wide measurement rather than a single-column anecdote.

\bb{Why not curate a fixed library once and skip the policy?} Because the answer is
unstable and cannot price optionality: the best fixed subset differs by benchmark ($T_4$
on GenBench-X, $T_3{+}T_4$ on BM4, $T_1{+}T_2{+}T_4$ on GenomeQA), and the tools a
fixed subset omits can still be the only ones that answer particular questions correctly
(Appendix~\ref{app:ceilings}).

\begin{figure}[h]
\centering
\includegraphics[width=0.94\linewidth]{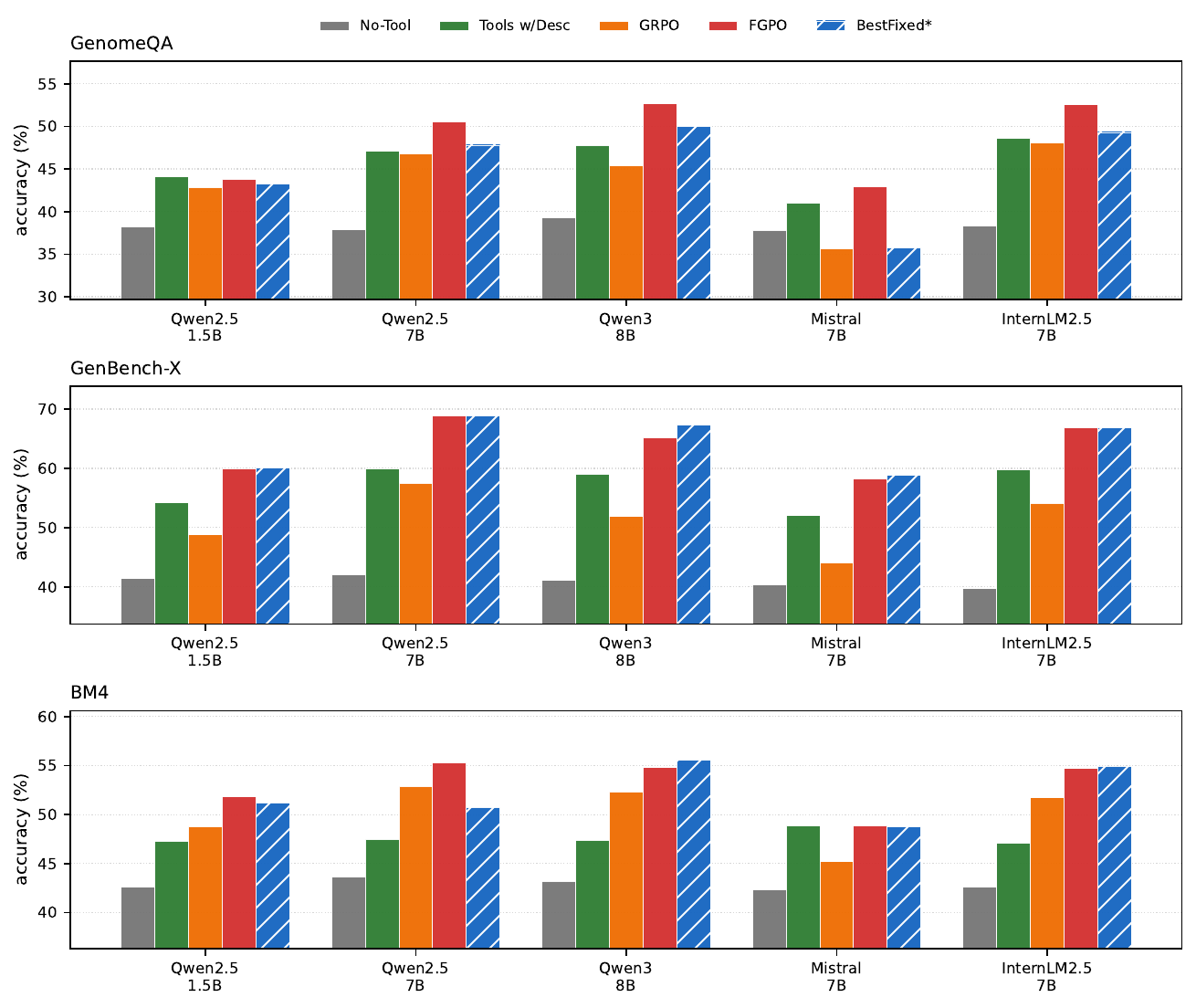}
\caption{Plug-in transfer per benchmark across five frozen reasoners (the grid of
Table~\ref{tab:main}). Random, SFT and DPO are omitted here for legibility and
reported in full in the table.}
\label{fig:transferbars}
\end{figure}

The numbers themselves are in Table~\ref{tab:main}; this section reads their shape.
Figure~\ref{fig:transferbars} plots the same grid per benchmark, which makes two things
visible that a table of $135$ numbers does not. First, more evidence is not
monotonically better: on Mistral-7B's GenomeQA column, invoking the whole library scores
$23.09$ against a no-tool floor of $37.83$, i.e.\ $14.7$ points of damage done purely by
handing the reasoner more to read.
Second, \method{} tracks BestFixed$^{*}$, a test-label-selected fixed reference that needs test labels to
pick its subset, closely enough that the gap is two-signed, which is what one expects when
a single fixed subset happens to be near-optimal for a given reasoner.

Figure~\ref{fig:deltacells} collapses the grid to the single comparison the paper is
built on. Every one of the $15$ cells favors \method{} over GRPO, and the spread is
informative rather than uniform: the margin is largest on GenBench-X ($+11.20$ to
$+14.20$), where tool evidence is most decisive, and smallest on GenomeQA with the $1.5$B
reasoner ($+0.92$), where the reasoner is too weak to exploit a better selection at all.
In this grid, variation across benchmarks exceeds variation across reasoners in the \method{}--GRPO margin.

\subsection{Pipeline agreement with the exhaustive table}
The GenomeQA cells of Table~\ref{tab:main} are produced by the transfer pipeline, which
calls the reasoner live, whereas Table~\ref{tab:economy} is read out of the exhaustive
cache. The two paths are independent implementations of the same measurement and agree to
within $0.05$ points on every row (\method{} $52.65$ vs.\ $52.62$; No-Tool $39.33$ vs.\
$39.28$; GRPO $45.43$ vs.\ $45.46$; All-Tools $49.44$ vs.\ $49.39$). A third path, the
per-epoch harness behind Tables~\ref{tab:kcurve} and~\ref{tab:datascale}, scores the same
\method{} checkpoint at $52.59$, so all three agree to within $0.06$ points. The residual is
response nondeterminism at the reasoner server, not a scoring difference.

\subsection{Ceilings: the best fixed subset and the per-question oracle}
\label{app:ceilings}
Two reference lines are computable from the exhaustive tables at no additional cost, and
we report them here rather than in the main comparison because neither is a deployable
method. BestFixed$^{*}$ applies one subset uniformly to every question: the best of all
$16$ chosen \emph{post-hoc on the test set} under the training reasoner Qwen3-8B, then
transplanted unchanged to the other four reasoners, exactly as the learned policy is. The
per-question oracle picks the best subset separately for each question. The oracle upper-bounds
every strategy; BestFixed$^{*}$ upper-bounds fixed subsets in the Qwen3-8B column and is a
test-label-selected fixed reference elsewhere, which is why it can fall below All-Tools or
No-Tool for a reasoner that ranks the subsets differently. Neither is a baseline a
practitioner could run.

Table~\ref{tab:ceilings} gives BestFixed$^{*}$ against \method{} on every cell.
\method{} leads it in $10$ of the $15$, and the remaining margins are $0.10$--$2.10$:
without access to test labels the learned policy closely tracks a reference selected with them. The
comparison also shows that the winning subset is not stable: it is
$T_4$ alone on GenBench-X, $T_3{+}T_4$ on BM4 and $T_1{+}T_2{+}T_4$ on
GenomeQA, so no single curation choice transfers across benchmarks.

\begin{table}[h]
\centering
\caption{\method{} against BestFixed$^{*}$, the fixed subset selected post-hoc on the
test set under Qwen3-8B and transferred to the other reasoners. Bold marks the larger of the
two per cell.}
\label{tab:ceilings}
\vspace{2pt}
{\footnotesize
\setlength{\tabcolsep}{4.6pt}
\begin{tabular}{ll ccccc}
\toprule
Benchmark & & Qwen3-8B & Qwen2.5-1.5B & Qwen2.5-7B & Mistral-7B & InternLM2.5-7B\\
\midrule
\multirow{2}{*}{GenomeQA}
 & \method{} & \best{52.65} & \best{43.79} & \best{50.58} & \best{42.90} & \best{52.53}\\
 & BestFixed$^{*}$ & 50.00 & 43.20 & 47.80 & 35.79 & 49.30\\
\midrule
\multirow{2}{*}{GenBench-X}
 & \method{} & 65.20 & 60.00 & \best{68.90} & 58.20 & \best{66.90}\\
 & BestFixed$^{*}$ & \best{67.30} & \best{60.10} & 68.80 & \best{58.80} & 66.80\\
\midrule
\multirow{2}{*}{BM4}
 & \method{} & 54.80 & \best{51.80} & \best{55.30} & \best{48.90} & 54.70\\
 & BestFixed$^{*}$ & \best{55.60} & 51.20 & 50.70 & 48.80 & \best{54.90}\\
\bottomrule
\end{tabular}}
\end{table}

The per-question oracle sits far above both: $77.41$ on GenomeQA under the training
reasoner, $90.60$ on GenBench-X and $86.90$ on BM4, against \method{}'s $52.62$, $65.20$
and $54.80$. We read that gap as a statement about the problem rather than about the
method (selection over these libraries is nowhere near solved) but it is an upper
bound no policy in this paper approaches, and we do not present it as one that any policy
should be expected to reach.

\subsection{Paired significance}
Question-paired McNemar tests, pooled per reasoner ($n\approx 5{,}590$):
\method{} vs.\ GRPO: $p = 2{\times}10^{-8}$ / $8{\times}10^{-19}$ / $2{\times}10^{-39}$ /
$2{\times}10^{-30}$ / $1{\times}10^{-20}$ (1.5B/7B/8B/Mistral/InternLM).
\method{} vs.\ Tools w/ Desc: $p = 2{\times}10^{-3}$ / $1{\times}10^{-16}$ /
$2{\times}10^{-18}$ / $2{\times}10^{-4}$ / $2{\times}10^{-15}$, a significant win in
every column.

\section{Additional Analyses}
\label{app:additional}
\subsection{\texorpdfstring{$k$}{k}-subset interpolation in full}
Table~\ref{tab:kcurve} gives every epoch of every $k$ arm; the main text reports peaks
only.

\begin{figure}[h]
\centering
\includegraphics[width=0.52\linewidth]{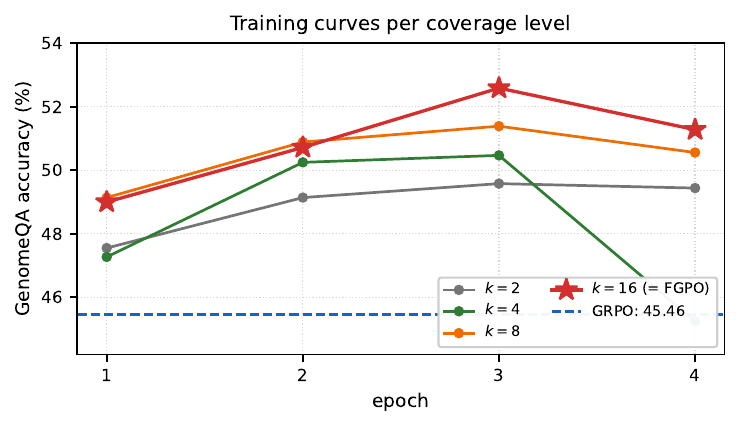}
\caption{The coverage dial: per-epoch training curves for $k\in\{2,4,8,16\}$ actions
evaluated per visit. $k{=}16$ is \method{} itself.}
\label{fig:kfull}
\end{figure}

Figure~\ref{fig:kfull} plots the same runs. Coverage does not merely shift the final
number: it shifts the whole trajectory, with every arm peaking at the dev-selected epoch 3
and the $k{=}4$ arm collapsing hardest afterwards. The peak values give the dose--response
the argument needs (Figure~\ref{fig:teaser}c): accuracy rises monotonically in $k$ while the
data, ordering, optimization settings and scoring construction are held fixed, so coverage
of the action space is doing the work rather than any of the confounds that separate
\method{} from GRPO.

\subsection{Per-question paired analysis and selection heatmap}
\label{app:paired}
\begin{figure}[h]
\centering
\includegraphics[width=\linewidth]{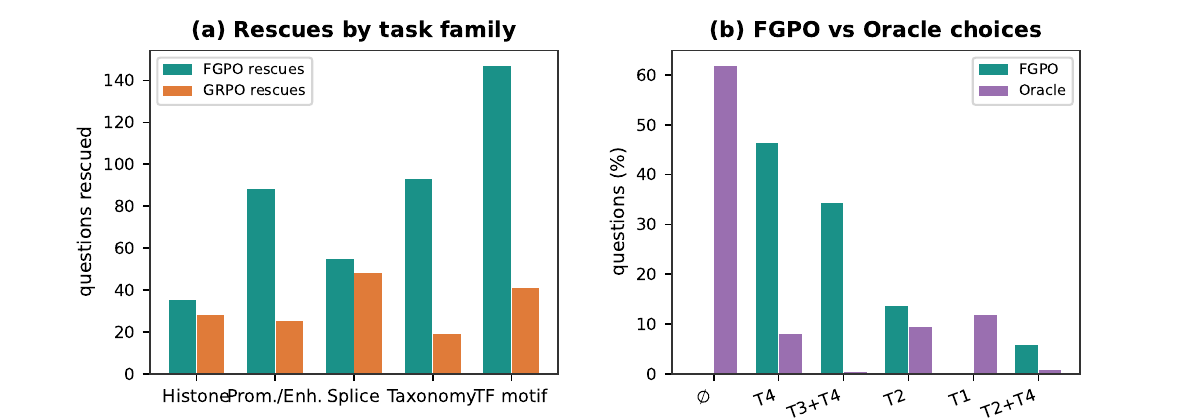}
\caption{\textbf{Per-question analysis on GenomeQA.} (a) Where the $418$ \method{} rescues
and the $161$ GRPO rescues of Figure~\ref{fig:analysis}c fall across task families.
(b) \method{} vs.\ the oracle: the oracle prefers the empty set on $61.9\%$ of questions,
yet \method{} almost never abstains.}
\label{fig:analysis_app}
\end{figure}

Figure~\ref{fig:analysis_app} decomposes the aggregate outcome of
Figure~\ref{fig:analysis}c by task family, and contrasts \method{}'s choices with the
per-question oracle.

\subsection{Data-scale ablation}
\label{app:datascale}
Table~\ref{tab:datascale} trains \method{} on subsets of the $2{,}002$-question training
pool. Accuracy is evaluated on the full $3{,}590$-question GenomeQA test set at every
epoch; the best epoch is dev-selected. With only $500$ training questions \method{} already
reaches $52.23$, within $0.4$ points of the full-data peak ($52.59$), and the $1{,}000$-question
arm peaks at $50.92$. The relationship is not monotone, since $500$ slightly outperforms
$1{,}000$, but all three are well above GRPO's $45.46$.

\begin{table}[h]
\centering
\caption{Data-scale ablation on GenomeQA. Accuracy (\%) at each epoch; peak is
dev-selected. All arms use the same hyperparameters.}
\label{tab:datascale}
\vspace{2pt}
{\small
\begin{tabular}{lccccc}
\toprule
Training $|\mathcal{D}|$ & ep1 & ep2 & ep3 & ep4 & peak\\
\midrule
$500$ & 46.99 & 50.03 & 51.14 & \best{52.23} & \best{52.23}\\
$1{,}000$ & 49.64 & 49.22 & \best{50.92} & 49.16 & 50.92\\
$2{,}002$ (full) & 49.00 & 50.72 & \best{52.59} & 51.28 & \best{52.59}\\
\midrule
GRPO ($2{,}002$) & \multicolumn{4}{c}{---} & 45.46\\
\bottomrule
\end{tabular}}
\end{table}

\subsection{Tool-library single-removal ablations}
Table~\ref{tab:libablation} prices each tool by what its removal costs the two ceilings,
computed from the exhaustive tables without running a model. No single tool is dispensable
on all three benchmarks, but the damage is wildly uneven: removing $T_4$ costs the
GenBench-X oracle $15.5$ points and the best fixed subset $20.5$, while removing $T_3$ costs
the best fixed subset at most $0.30$ anywhere, an option that only pays on questions a
fixed subset was never going to get right.

\begin{table}[h]
\centering
\caption{Ceiling loss from removing one tool from the full library (computed from the
exhaustive tables; no model runs). $\Delta$oracle / $\Delta$best-fixed in points.}
\label{tab:libablation}
\vspace{2pt}
{\scriptsize
\setlength{\tabcolsep}{3.6pt}
\begin{tabular}{lcccc}
\toprule
Benchmark & $-T_1$ & $-T_2$ & $-T_3$ & $-T_4$\\
\midrule
GenomeQA & $-3.29$ / $0.00$ & $-6.21$ / $-2.20$ & $-3.26$ / $0.00$ & $-10.84$ / $-4.76$\\
GenBench-X & $-2.10$ / $0.00$ & $-2.70$ / $0.00$ & $-1.40$ / $0.00$ & $-15.50$ / $-20.50$\\
BM4 & $-2.70$ / $0.00$ & $-3.10$ / $0.00$ & $-1.20$ / $-0.30$ & $-14.40$ / $-10.00$\\
\bottomrule
\end{tabular}}
\end{table}

\section{Baseline Fairness Sweeps}
\label{app:fairness}
No baseline in this paper is reported at a single arbitrary configuration.
\subsection{GRPO checkpoint and configuration sweeps}
\begin{table}[h]
\centering
\caption{GRPO on GenomeQA: all eight checkpoints of the reported run, and four further
independent configurations. The reported $45.46$ is the test accuracy at the dev-selected
checkpoint (step~75).}
\vspace{2pt}
{\small
\begin{tabular}{lcccccccc}
\toprule
Checkpoint & 25 & 50 & \textbf{75} & 100 & 125 & 150 & 175 & 200\\
\midrule
Accuracy & 42.31 & 44.85 & \best{45.46} & 45.24 & 44.90 & 45.32 & 45.29 & 45.13\\
Mean tools & 1.33 & 1.92 & 2.36 & 2.18 & 2.17 & 2.25 & 2.15 & 2.17\\
\midrule
\multicolumn{9}{l}{\scriptsize Independent configurations (varying $G$, temperature, data mix): $43.90$ / $44.26$ / $44.35$ / $45.57$.}\\
\bottomrule
\end{tabular}}
\end{table}
\subsection{DPO \texorpdfstring{$\beta$}{beta} sweep}
\begin{table}[h]
\centering
\caption{DPO on GenomeQA: every checkpoint of every $\beta$. The best observed cell
reaches $39.33$, essentially the no-tool floor.}
\vspace{2pt}
{\small
\begin{tabular}{lccc}
\toprule
$\beta$ & Checkpoint accuracies & Mean tools & Verdict\\
\midrule
0.1 & 39.25 / 39.28 / 39.25 & 0.00 & collapsed\\
0.3 & 39.28 / 39.33 & 0.03 / 0.01 & collapsed\\
0.5 & 38.97 / 39.00 & 0.10 / 0.08 & collapsed\\
\bottomrule
\end{tabular}}
\end{table}

\begin{figure}[h]
\centering
\includegraphics[width=0.96\linewidth]{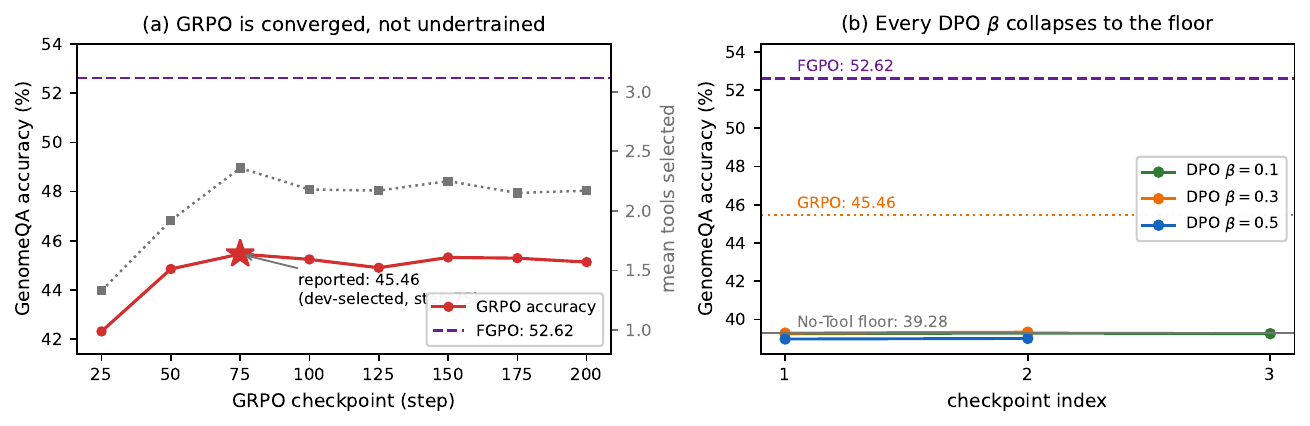}
\caption{Baseline fairness sweeps on GenomeQA. (a) GRPO across all eight checkpoints,
mean tools on the right axis. (b) Every checkpoint of every DPO $\beta$.}
\label{fig:sweeps}
\end{figure}

Figure~\ref{fig:sweeps} shows both sweeps. Panel (a) is the check that matters for the
headline comparison: GRPO's accuracy is flat from step $50$ onwards while its mean tool
count is also flat, so the reported $45.46$ is a converged plateau rather than an
undertrained checkpoint we happened to stop at. Panel (b) shows the DPO failure has no
$\beta$ that rescues it: every checkpoint of every $\beta$ lands within $0.4$ points of
the no-tool floor, because the policy has stopped selecting tools at all.

\section{Mechanism Details}
\label{app:mechanism}

\subsection{Two properties of dead groups and of the exact objective}
\label{app:theory}
Fix a question $s$, its reward table $r(s,\cdot)$ over $\mathcal{A}$, and a sampling
distribution $p$ over $\mathcal{A}$ (for GRPO, $p=p_\theta^{\mathrm{gen}}(\cdot\,|\,s)$).
Partition $\mathcal{A}$ into reward classes $\{C_v\}$, one per distinct reward value $v$, and
write $w_v=\sum_{a\in C_v}p(a)$ for the class masses, so $\sum_v w_v=1$. A group of $G$
i.i.d.\ draws is dead when all $G$ land in one class, which happens with probability
\begin{equation}
P_{\mathrm{dead}}(p,G)\;=\;\sum_{v} w_v^{G}.
\label{eq:pdead}
\end{equation}

\paragraph{Proposition 1 (reward collisions).} Let $G\ge 2$ be an integer. (i) \emph{Coarsening.} If two
reward classes are merged (as happens when a finer reward is replaced by a coarser one that
assigns them the same value), $P_{\mathrm{dead}}$ does not decrease. (ii)
\emph{Concentration.} If mass $0<\delta\le w_u$ is moved from a class $u$ to a distinct
class $v$ with $w_v\ge w_u$, leaving all other masses fixed, $P_{\mathrm{dead}}$ strictly
increases. For a fixed number of classes, $P_{\mathrm{dead}}$ is Schur-convex in $(w_v)_v$, so it is
minimized at the uniform class distribution and maximized when one class carries all the
mass.

\emph{Proof.} (i) For $a,b\ge 0$ and $G\ge 2$, $(a+b)^{G}=\sum_{j=0}^{G}\binom{G}{j}a^{j}b^{G-j}\ge a^{G}+b^{G}$,
so replacing the two terms $w_u^{G}+w_v^{G}$ in Eq.~\ref{eq:pdead} by $(w_u+w_v)^{G}$ cannot
lower the sum. (ii) The function $f(x)=x^{G}$ is strictly convex on $[0,\infty)$ for
$G\ge 2$, so $f(w_v+\delta)+f(w_u-\delta)-f(w_v)-f(w_u)
=\int_{0}^{\delta}\big[f'(w_v+t)-f'(w_u-t)\big]\,dt>0$ because $f'$ is strictly increasing and
$w_v+t>w_u-t$ for all $t\in(0,\delta]$. Schur-convexity follows because
$\sum_v f(w_v)$ with $f$ convex is Schur-convex \citep{marshall1979inequalities}. $\square$

Proposition 1 gives conditions for interpreting the diagnostics of
\secref{sec:exp:mechanism}. Under the same sampling distribution, part (i) orders the
$20.8\%$ closed-form rate under the fine partition of Eq.~\ref{eq:reward} below the
$79.6\%$ rate under the coarser differential reward. Part (ii) concerns concentration of
\emph{reward-class mass}; a decrease in policy entropy alone does not establish this
condition. The concurrent entropy decrease and dead-rate increase in
Figure~\ref{fig:dynamics} are consistent with this mechanism, but do not establish a
majorization ordering of the class-mass vectors. The propositions apply regardless of
whether the policy is correct; the observed near-independence of dead mass and correctness
is an empirical finding of \secref{sec:exp:mechanism}, not a consequence of these properties.

\paragraph{Proposition 2 (the exact objective).} Fix $s$ and $\theta$, and let
$q_\theta(\cdot\,|\,s)$ be the softmax in Eq.~\ref{eq:score} with finite logits
$z_a=\ell_\theta(a\,|\,s)$. Let
$J_s(\theta)=\sum_{a}q_\theta(a\,|\,s)\,r(s,a)$ be the per-question objective of
Eq.~\ref{eq:exact}. In exact arithmetic, (i) $J_s$ and $\nabla_\theta J_s$ are computed exactly from the
$|\mathcal{A}|$ candidate scores, so the estimator has zero variance with respect to action
sampling. Random minibatches can still introduce variance across questions.
(ii) The reward-term gradient with respect to the logit of action $a$ is
\begin{equation}
\frac{\partial J_s}{\partial z_a}\;=\;q_\theta(a\,|\,s)\,\big(r(s,a)-J_s\big),
\label{eq:logitgrad}
\end{equation}
Since finite softmax logits give $q_\theta(a\,|\,s)>0$, every action with $r(s,a)\ne J_s$
has a non-zero reward-term logit gradient whose sign is that of its
advantage over the current expected reward, and whose magnitude is proportional to
$q_\theta(a\,|\,s)$.

\emph{Proof.} (i) is immediate from Eq.~\ref{eq:exact} being a finite sum with no sampled
term. (ii) With $q_a=\exp z_a/\sum_{a'}\exp z_{a'}$, $\partial q_b/\partial z_a
= q_b(\mathbf{1}[a{=}b]-q_a)$, hence $\partial J_s/\partial z_a
=\sum_b r_b\,q_b(\mathbf{1}[a{=}b]-q_a)= q_a r_a - q_a\sum_b q_b r_b = q_a(r_a-J_s)$. $\square$

These statements concern the reward term in logit space. Exhaustive evaluation rules out
zeros caused by sampled reward collisions, but small $q_\theta(a\,|\,s)$ can still make
the logit gradient arbitrarily small. LoRA parameter updates couple candidate logits, so a
non-zero logit derivative guarantees neither an increase in that action's probability nor
fast escape. The entropy bonus of Eq.~\ref{eq:entropy} adds a separate gradient and
encourages dispersion; it guarantees neither a probability floor during training nor rapid
recovery. The $k$-subset intervention of \secref{sec:exp:kcurve} changes action coverage
and renormalizes $q_\theta$ within each sampled subset, thereby changing the objective;
its gradient need not be unbiased for $J_s$. At $k=|\mathcal{A}|$, it recovers the full
exact objective.

\subsection{Dead-group grids for all three benchmarks}
Table~\ref{tab:deadgroups} gives the closed-form rates behind
Figure~\ref{fig:deadpanels} at every group size we evaluated, for all four exported
distributions rather than the two the main text plots.
\begin{table}[h]
\centering
\caption{Closed-form dead-group rate (\%) vs.\ group size $G$, per benchmark, the
numerical form of Figure~\ref{fig:deadpanels}a--c, computed from each policy's
generation-induced distribution over the $16$ valid parsed subset strings.
Reward classes follow Eq.~\ref{eq:reward}, the finest partition any arm induces. SFT is
reported on GenomeQA only; GRPO and \method{} are reported on all three benchmarks.}
\label{tab:deadgroups}
\vspace{2pt}
{\scriptsize
\setlength{\tabcolsep}{3.6pt}
\begin{tabular}{llccccccc}
\toprule
Benchmark & Policy & $G{=}1$ & $G{=}2$ & $G{=}4$ & $G{=}6$ & $G{=}8$ & $G{=}16$ & $G{=}32$\\
\midrule
GenomeQA & Uniform & 100.0 & 22.7 & 1.9 & 0.2 & 0.0 & 0.0 & 0.0\\
 & SFT & 100.0 & 40.4 & 9.1 & 2.4 & 0.8 & 0.0 & 0.0\\
 & GRPO & 100.0 & 56.5 & 30.4 & \bb{20.8} & 15.7 & 7.7 & 3.7\\
 & \method{} & 100.0 & 97.3 & 95.1 & 94.0 & 93.2 & 91.5 & 89.7\\
\midrule
GenBench-X & Uniform & 100.0 & 20.4 & 1.4 & 0.1 & 0.0 & 0.0 & 0.0\\
 & GRPO & 100.0 & 53.5 & 26.9 & \bb{17.8} & 13.4 & 6.2 & 1.8\\
 & \method{} & 100.0 & 97.1 & 94.7 & 93.4 & 92.5 & 90.5 & 88.7\\
\midrule
BM4 & Uniform & 100.0 & 20.5 & 1.5 & 0.1 & 0.0 & 0.0 & 0.0\\
 & GRPO & 100.0 & 56.1 & 29.0 & \bb{17.4} & 11.1 & 2.9 & 0.8\\
 & \method{} & 100.0 & 98.4 & 97.2 & 96.6 & 96.2 & 95.4 & 94.7\\
\bottomrule
\end{tabular}}
\end{table}

Read across the three grids, the two curves that matter run in opposite directions and
neither is an accident of tuning: a uniform reference policy is almost never dead beyond
$G{=}4$, and under the fine-grained Eq.~\ref{eq:reward} partition the converged GRPO policy
has a $17$--$21\%$ dead-group probability at the $G{=}6$ group size used for training.
\method{}'s own row is included precisely because it is the worst of the four: its converged
distribution would be nearly useless to sample from ($94\%$ dead at $G{=}6$), which is the
point, since it never samples. SFT's low dead rate is not evidence of useful routing: its
sampling distribution stays relatively diffuse, but its argmax collapses onto the empty
action on $92.8\%$ of questions, and its accuracy ($42.17$) trails GRPO's regardless.

\subsection{Empty-subset preference: how offline objectives collapse}
\begin{table}[h]
\centering
\caption{Generation-induced probability mass and argmax rate on the empty subset (no
tools), per trained policy, over the GenomeQA test set; both columns are computed from
$p_\theta^{\mathrm{gen}}$, the same distribution as Table~\ref{tab:deadgroups}. Reference:
the empty subset is reward-optimal on $61.9\%$ of questions.}
\vspace{2pt}
{\small
\begin{tabular}{lccc}
\toprule
Policy & Generation-induced mass on $\emptyset$ & Argmax $=\emptyset$ & GenomeQA accuracy\\
\midrule
SFT & 42.2\% & 92.8\% & 42.17\\
GRPO & 4.0\% & 2.8\% & 45.46\\
\method{} & 0.2\% & 0.2\% & 52.62\\
\bottomrule
\end{tabular}}
\end{table}
SFT's high mass on the modal action also explains its deceptively favorable
``chance-of-sampling-an-optimal-action'' statistics: mimicking the mode looks good on any
metric that ignores the other $38\%$ of questions.

\begin{figure}[h]
\centering
\includegraphics[width=0.55\linewidth]{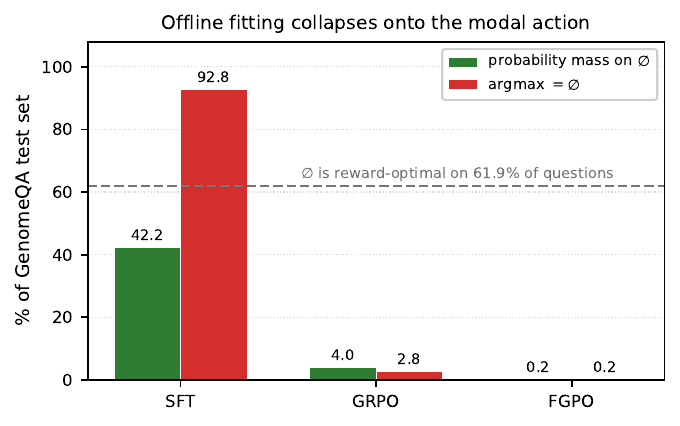}
\caption{Generation-induced probability mass and argmax rate on the empty subset per
trained policy. The empty subset is reward-optimal on $61.9\%$ of GenomeQA test questions
(dashed line).}
\label{fig:libempty}
\end{figure}

Figure~\ref{fig:libempty} shows the mechanism behind the offline collapse: the empty subset
is reward-optimal on $61.9\%$ of questions, SFT puts $92.8\%$ of its argmax there, and the
resulting policy is a very good imitation of the modal answer and a very poor policy.

\section{Prompt Templates}
\label{app:prompts}
\subsection{Policy prompt (all trained arms: \method{}, GRPO, SFT, DPO)}
Tools are anonymous slots; the policy answers in indices. System:
\begin{quote}\ttfamily\small
You are an expert agent specialized in selecting tools to solve genomic reasoning tasks.
You are provided with access to \{n\} tools, indexed from 0 to \{n-1\}.
\end{quote}
Tool list and instruction:
\begin{quote}\ttfamily\small
0: type1 (A)\\
1: type2 (B)\\
2: type3 (C)\\
3: type4 (D)\\[2pt]
\{question and options\}\\[2pt]
Select the index number(s) of the tools that are most helpful for solving the task. You
MUST output only the selected tool indices as a comma-separated list, enclosed in
<answer></answer> tags. Output an empty <answer></answer> if no tool is needed.
\end{quote}
An empty \texttt{<answer></answer>} selects no tools; our action space includes it
explicitly.

\subsection{Prompted baseline (Tools w/ Desc)}
Identical layout, but each slot carries the tool's real name and a functional
description:
\begin{quote}\ttfamily\small
\{i\}: \{tool name\} - Description: \{functional description\}
\end{quote}
The prompted baseline therefore receives strictly more prior information than any
trained policy in this paper.

\subsection{Reasoner prompt}
The frozen reasoner receives the question, the options, and the rendered evidence of the
selected subset (one titled block per tool, shared verbatim across all methods and table
rows), and is instructed to answer with a single option letter.

\subsection{Complete reasoner prompt}
\label{app:reasoner_full}
The frozen reasoner receives the following system message and user prompt. The evidence
block is empty in the No-Tool arm and contains one titled block per selected tool in all
other arms. The same template is shared across every method and every table row; only
the evidence differs.

\paragraph{System message.}
\begin{quote}\ttfamily\small
You are a genomics expert answering multiple-choice questions about DNA sequences. Tool
outputs, when provided, may be incomplete or noisy. Use only the evidence that is relevant
to the question and ignore the rest. Do not assume a tool is correct. Respond with the
single letter of the best option and nothing else.
\end{quote}

\paragraph{User message template.}
\begin{quote}\ttfamily\small
\{question\}\\[4pt]
Options:\\
\{options\_block\}\\[4pt]
Analysis tool output for the sequence above:\\
\{evidence\}\\[4pt]
Answer with a single letter.
\end{quote}
When no tools are selected, the evidence block is omitted entirely.

\subsection{Complete tool evidence examples}
\label{app:evidence_examples}
Below are schematic evidence blocks condensed from the exhaustive reward tables; the prompt itself carries each tool's JSON output in full.
Each example illustrates the output format of one tool; the reasoner sees the concatenation
of the selected tools' blocks.

\paragraph{$T_1$ (Sequence Composition).}
\begin{quote}\ttfamily\small
Sequence A: length=371bp, GC=59.3\%, AT=40.7\%, CpG O/E=0.264, GC\_skew=0.051\\
Top 3-mers: TTT (12, 3.3\%), TAT (9, 2.4\%), TTA (8, 2.2\%)\\
Sequence B: length=371bp, GC=42.1\%, AT=57.9\%, CpG O/E=0.891, GC\_skew=-0.032\\
Top 3-mers: AGG (11, 3.0\%), GGA (9, 2.4\%), CAG (8, 2.2\%)\\
\ldots (C, D analogous)
\end{quote}

\paragraph{$T_2$ (Motif Scanner).}
\begin{quote}\ttfamily\small
Sequence A: JASPAR hits (p<1e-4): JUN (score=12.3), FOS (11.8), TEAD4 (10.2)\\
Sequence B: STAT1 (score=14.1), IRF1 (9.7)\\
Sequence C: SP1 (score=13.5), KLF4 (12.1)\\
Sequence D: SRF (score=15.8), MEF2A (11.2)
\end{quote}

\paragraph{$T_3$ (Splice-Site Analysis).}
\begin{quote}\ttfamily\small
Sequence A: no GT/AG sites above threshold (donor>6.0, acceptor>6.0)\\
Sequence B: acceptor at pos 142 (score=8.53, Only Acceptor)\\
Sequence C: no sites above threshold\\
Sequence D: no sites above threshold
\end{quote}

\paragraph{$T_4$ (Genomic Expert).}
\begin{quote}\ttfamily\small
$k$NN prediction ($k{=}15$, cosine over frozen NT-v2-50m embeddings; confidence is the\\
distance-weighted vote share, counts are raw neighbours):\\
Sequence A: Eukaryote 0.82 (12/15 neighbours)\\
Sequence B: Virus 0.37 (6/15 neighbours; Eukaryote 0.33, Prokaryote 0.30)\\
Sequence C: Eukaryote 0.53 (8/15)\\
Sequence D: Eukaryote 0.84 (13/15)
\end{quote}

\paragraph{$T_5$ (Homology Search), used only in the five-tool stress test of
\secref{app:fivetool}.}
\begin{quote}\ttfamily\small
Sequence A: 2 hits. ref\_04188 (label=promoter, identity=71.4\%, coverage=58\%);\\
\phantom{Sequence A: }ref\_11902 (label=enhancer, identity=64.0\%, coverage=41\%).\\
\phantom{Sequence A: }Consensus: promoter (1/2, agreement 0.50)\\
Sequence B: no hit above threshold (identity $\geq 60\%$, coverage $\geq 30\%$)\\
Sequence C: 1 hit. ref\_00734 (label=enhancer, identity=62.8\%, coverage=35\%).\\
\phantom{Sequence C: }Consensus: enhancer (1/1, agreement 1.00)\\
Sequence D: no hit above threshold
\end{quote}
Two of the four candidates return nothing; among the remaining two, one has only a $1/2$
consensus and the other rests on a single supporting hit. Evidence of this shape is why the exact objective settles on invoking $T_5$ for only
$2.8\%$ of questions (\secref{app:fivetool}): the tool is well implemented but rarely
decisive on this benchmark, and the reward makes that visible without any hand-written rule.

\subsection{Question format examples}
\label{app:question_examples}
Table~\ref{tab:qexamples} shows one verbatim question from each benchmark, illustrating the
diversity of task types, option formats and sequence lengths. Every option list is
reproduced in full; only the nucleotide sequence is elided in the middle.

\begin{table}[h]
\centering
\caption{\textbf{Question format across the three benchmarks.} One row per benchmark, quoted
verbatim from the released files. \textcolor{gtgreen}{Green} marks the gold option.
Sequences are shown head\,$\ldots$\,tail; the full length is given in the header row.}
\label{tab:qexamples}
\vspace{3pt}
{\footnotesize
\setlength{\tabcolsep}{3.5pt}
\renewcommand{\arraystretch}{1.12}
\begin{tabular}{@{}p{0.30\linewidth}|p{0.64\linewidth}@{}}
\toprule
\multicolumn{1}{c|}{\textbf{Question}} & \multicolumn{1}{c}{\textbf{Options}}\\
\midrule
\rowcolor{grouprow}[0pt][0pt]\multicolumn{2}{@{}p{0.985\linewidth}@{}}{\textbf{GenomeQA}
--- TF motif, \texttt{describe} shape, $100$\,bp}\\
``Identify the transcription factor that binds to this Human DNA sequence:
\texttt{CCCTACAGAG\ldots CTCGAGGCTC}'' &
\texttt{A: NRF1} \quad \texttt{B: POU5F1} \quad \texttt{C: E2F1} \quad
\textcolor{gtgreen}{\texttt{\textbf{D: RELA}}}\\
\midrule
\rowcolor{grouprow}[0pt][0pt]\multicolumn{2}{@{}p{0.985\linewidth}@{}}{\textbf{GenBench-X}
--- taxonomy, \texttt{describe} shape, $200$\,bp}\\
``Which organism is this DNA sequence from?''
\texttt{CCCAGCCCTA\ldots GTCAGACAGT} &
\texttt{A: Caenorhabditis elegans (worm)} \quad
\textcolor{gtgreen}{\texttt{\textbf{B: Human}}}\\
\midrule
\rowcolor{grouprow}[0pt][0pt]\multicolumn{2}{@{}p{0.985\linewidth}@{}}{\textbf{BM4}
--- bacterial promoter, \texttt{describe} shape, $81$\,bp}\\
``Is this \emph{Escherichia coli} DNA sequence a sigma70 promoter?''
\texttt{TCTACGCCGC\ldots AGGTAGAAGT} &
\textcolor{gtgreen}{\texttt{\textbf{A: Not a promoter}}} \quad
\texttt{B: Promoter}\\
\bottomrule
\end{tabular}}
\end{table}

\subsection{Tool description prompt (Tools w/ Desc baseline)}
\label{app:desc_prompt}
The prompted baseline receives the same layout as the trained policies, but each tool
slot carries the real name and a functional description:
\begin{quote}\ttfamily\small
0: sequence\_statistics - Computes GC content, CpG density, $k$-mer frequencies, and
   other compositional statistics of the DNA sequence.\\[2pt]
1: motif\_scanner - Scans the sequence against JASPAR transcription factor binding
   profiles and returns significant hits with their scores.\\[2pt]
2: splice\_analyzer - Identifies candidate GT/AG splice sites and scores them with
   MaxEntScan donor and acceptor models.\\[2pt]
3: genomic\_expert - Classifies the sequence by $k$NN vote over frozen
   Nucleotide-Transformer embeddings of a labelled reference set.
\end{quote}
The prompted baseline therefore receives strictly more prior information than any trained
policy in this paper. Its accuracy ($47.77$ on GenomeQA) exceeds GRPO ($45.43$), showing
that access to tool semantics is valuable, but \method{} surpasses both ($52.65$) without
any semantic information.

\end{document}